\documentclass{article}
\usepackage{arxiv}
\usepackage{authblk}
\usepackage[utf8]{inputenc} 
\usepackage[T1]{fontenc}    
\usepackage{hyperref}       
\usepackage{url}            
\usepackage{booktabs}       
\usepackage{amsmath}
\usepackage{amsfonts}       
\usepackage{nicefrac}       
\usepackage{microtype}      
\usepackage{amssymb}
\usepackage{systeme}
\usepackage{bbm}
\usepackage{stmaryrd}
\newcommand{\ssymbol}[1]{^{\@fnsymbol{#1}}}
\usepackage{bm}
\usepackage{graphicx}
\usepackage{natbib}
\usepackage{doi}
\usepackage{subfig}
\usepackage{float}
\usepackage{multirow}

\usepackage{algorithm}
\usepackage{algorithmic}

\makeatletter
\renewcommand\AB@authnote[1]{\textsuperscript{\normalfont\bfseries#1}}
\makeatother

\title{Fixed-Budget Online Adaptive Learning for Physics-Informed Neural Networks.\\ Towards Parameterized Problem Inference.}

\author[1,2,3]{Thi Nguyen Khoa Nguyen}
\author[2]{Thibault Dairay}
\author[2]{Raphaël Meunier}
\author[1,3]{Christophe Millet}
\author[1,4]{Mathilde Mougeot}
\affil[1]{Universite Paris-Saclay, ENS Paris-Saclay, CNRS, Centre Borelli, Gif-sur-Yvette, 91190 France}
\affil[2]{Michelin, Centre de Recherche de Ladoux, Cébazat, 63118 France}
\affil[3]{CEA, DAM, DIF, F-91297 Arpajon, France}
\affil[4]{ENSIIE, Évry-Courcouronnes, 91000 France}
\begin{document}
\maketitle

\begin{abstract}
Physics-Informed Neural Networks (PINNs) have gained much attention in various fields of engineering thanks to their capability of incorporating physical laws into the models. PINNs integrate the physical constraints by minimizing the partial differential equations (PDEs) residuals on a set of collocation points. The distribution of these collocation points appears to have a huge impact on the performance of PINNs and the assessment of the sampling methods for these points is still an active topic. In this paper, we propose a Fixed-Budget Online Adaptive Learning (FBOAL) method, which decomposes the domain into sub-domains, for training collocation points based on local maxima and local minima of the PDEs residuals. The effectiveness of FBOAL is demonstrated for non-parameterized and parameterized problems. The comparison with other adaptive sampling methods is also illustrated. The numerical results demonstrate important gains in terms of the accuracy and computational cost of PINNs with FBOAL over the classical PINNs with non-adaptive collocation points. We also apply FBOAL in a complex industrial application involving coupling between mechanical and thermal fields. We show that FBOAL is able to identify the high-gradient locations and even give better predictions for some physical fields than the classical PINNs with collocation points sampled on a pre-adapted finite element mesh built thanks to numerical expert knowledge. From the present study, it is expected that the use of FBOAL will help to improve the conventional numerical solver in the construction of the mesh.
\end{abstract}



\section{Introduction}
In the last few years, Physics-Informed Neural Networks (PINNs) \citep{raissi2019physics} rise as an attractive and remarkable scheme of solving inverse and ill-posed partial differential equations (PDEs) problems. During the training process, PINNs aim to minimize (i) the loss on the initial/boundary conditions if the problem is well-defined, (ii) the loss on supervised data (if available), and (iii) the loss of PDE residuals on collocation points (unsupervised points). Thanks to their simplicity and powerful capability when dealing with forward or inverse problems, PINNs are gaining more and more attention from researchers and they are gradually being improved and extended aa. Furthermore, the applicability of PINNs has been demonstrated in various fields of research and industrial applications. For example, \cite{kissas2020machine} employed PINNs for the modeling of cardiovascular flows, \cite{chen2020physics} applied PINNs for inverse problems in nano-optics and metamaterials, \cite{shukla2021physics} used PINNs to identify polycrystalline nickel material coefficients, \cite{nguyen2022physics} applied PINNs in a non-Newtonian fluid thermo-mechanical problem which is used in the rubber calendering process.

As PINNs integrate the PDEs constraints by minimizing the PDE residuals on a set of collocation points during the training process, it has been shown that the location of these collocation points has a great impact on the performance of PINNs \citep{daw2022rethinking, nguyen2022physics}. To the best of the authors knowledge, the first work that showed the improvement of PINNs performance by modifying the set of collocation points is introduced by \cite{lu2021deepxde}. This work proposed the Residual-based Adaptive Refinement (RAR) that adds new training collocation points to the location where the PDE residual errors are large. RAR has been proven to be very efficient to increase the accuracy of the prediction \citep{yu2022gradient, hanna2022residual} but however leads to an uncontrollable amount of collocation points and computational cost at the end of the training process. \cite{nguyen2022physics} demonstrated in their use case that the performance of PINNs can be significantly improved by taking collocation points on a finite-element (FE) mesh, that provides an \textit{a priori} knowledge of high gradient location. However, this approach requires prior expertise on the mesh which is not always available.

In this work, we propose a Fixed-Budget Online Adaptive Learning (FBOAL) that fixes the number of collocation points during the training. The method adds and removes the collocation points based on the PDEs residuals on sub-domains during the training. By dividing the domain into smaller sub-domain it is expected that local maxima and minima of the PDEs residuals will be quickly captured by the method. Furthermore, the stopping criterion is chosen based on a set of reference solutions, which leads to an adaptive number of iterations for each specific problem and thus avoids unnecessary training iterations. We first show the potential of this method in the problem of solving Burgers equation in a setup where the viscosity is fixed (which we refer to as a non-parameterized problem) and in a setup where the viscosity is varied (which we refer to as parameterized problem). Another test on a wave equation is provided. The numerical results demonstrate that the use of FBOAL help to reduce remarkably the computational cost and gain significant accuracy compared to the conventional method of non-adaptive training points. We then apply this method to a non-Newtonian fluid thermo-mechanical problem, which is commonly used to model rubber manufacturing processes in the tire industry. We show that FBOAL is able to not only identify the high-gradient locations as provided by a finite-element mesh built thanks to expert knowledge but also point out the zones of contact with the rotating cylinders, which are not previously captured. 


In the very last months, several works have also introduced a similar idea of adaptive re-sampling of the PDE residual points during the training \citep{peng2022rang,daw2022rethinking, zeng2022adaptive,wu2023comprehensive}. \cite{wu2023comprehensive} give an excellent general review of these methods and proposed two adaptive re-sampling methods named Residual-based Adaptive Distribution (RAD) and Residual-based adaptive refinement with distribution (RAR-D). The authors also demonstrate that all existing methods of adaptive re-sampling for the collocation points are special cases (or with minor modifications) of RAD and RAR-D. These approaches aim to minimize the PDEs residuals at their global maxima on the entire domain. Besides that, the existing studies did not investigate the parameterized PDE problems (where the parameter of interest is varied). In this study, we first compare the performance of RAD, RAR-D, and FBOAL in the academic test cases (Burgers equation, wave equation). We illustrate a novel utilization of these adaptive sampling methods in the context of parameterized problems.

The following of this paper is organized as follows. In section 2, we briefly review the framework of PINNs and introduce the adaptive learning strategy (FBOAL) for the collocation points. We then provide the numerical results of the performance of the studied methods and comparison to the classical PINNs and other adaptive re-sampling methods such as RAD and RAR-D in a test case of Burgers equation. The application to an industrial use case is also represented in this section. Finally, we summarize the conclusions in section 4. The complement results on other test cases are provided in the appendix.

\section{Methodology}
In this section, the framework of Physics-Informed Neural Networks (PINNs) \citep{raissi2019physics} is briefly presented. Later, Fixed-Budget Online Adaptive Learning (FBOAL) for PDE residual points is introduced.

\subsection{Physics-informed neural networks}
To illustrate the methodology of PINNs, let us consider the following parameterized PDE defined on the domain $\Omega \subset \mathbb{R}^d$ with the boundary $\partial\Omega$:
\begin{align*}
    \bm{u}_t + \mathcal{N}_{\bm{\mathrm{x}}}(\bm{u}, \bm{\lambda}) = 0, \text{ for } \bm{\mathrm{x}}\in \Omega, t\in [0, T];\quad
    \mathcal{B}(\bm{u},\bm{\mathrm{x}},t) = 0, \text{ for } \bm{\mathrm{x}} \in \partial \Omega; \quad
    \bm{u}(\bm{\mathrm{x}}, 0) = g(\bm{\mathrm{x}}), \text{ for } \bm{\mathrm{x}}\in \Omega
\end{align*}
where $\bm{\mathrm{x}} \in \mathbb{R}^d$ and $t$ are the spatial and temporal coordinates, $\mathcal{N}_{\bm{\mathrm{x}}}$ is a differential operator, $\bm{\lambda} \in \mathbb{R}^p$ is the PDE parameter, $\bm{u}$ is the solution of the PDE with initial condition $g(\bm{\mathrm{x}})$ and boundary condition $\mathcal{B}$. The subscripts denote the partial differentiation in time or space. In the conventional framework of PINNs, the solution $\bm{u}$ of the PDE is approximated by a fully-connected feed-forward neural network $\mathcal{NN}$ and the prediction for the solution can be represented as $\bm{\hat{u}} = \mathcal{NN}(\bm{\mathrm{x}}, t, \bm{\theta})$ where $\bm{\theta}$ denotes the trainable parameters of the neural network. The parameters of the neural network are trained by minimizing the cost function $L = L_{pde} + w_{ic}L_{ic} + w_{bc}L_{bc}$, where the terms $L_{pde},L_{ic},L_{bc}$ penalize the loss in the residual of the PDE, the initial condition, the boundary condition, respectively, and $w_{ic}, w_{bc}$ are the positive weight coefficients to  adjust the contribution of each term to the cost function:
\begin{align*}
    L_{pde} = \dfrac{1}{N_{pde}}\sum_{i=1}^{N_{pde}}|\hat{\bm{u}}_{t_i} + \mathcal{N}_{\bm{\mathrm{x_i}}}(\hat{\bm{u}}, \bm{\lambda})|^2; \quad
    L_{ic} = \dfrac{w_{ic}}{N_{ic}}\sum_{i=1}^{N_{ic}}|\bm{\hat{u}}(\bm{\mathrm{x}}_i,0) - g(\bm{\mathrm{x}}_i)|^2; \quad
    L_{bc} = \dfrac{w_{bc}}{N_{bc}}\sum_{i=1}^{N_{bc}}|\mathcal{B}(\bm{\hat{u}}, \bm{\mathrm{x_i}}, t_i)|^2
\end{align*}
where $N_{ic}, N_{bc}, N_{data}$ denote the numbers of learning points for the initial condition, boundary condition, and measurements (if available), respectively, and $N_{pde}$ denotes the number of residual points (or collocation points or unsupervised points) of the PDE.

We note that PINNs may provide different performances with different network initialization. In this work when comparing the results of different configurations of PINNs, we train PINNs five times and choose the mean and standard deviation values of the performance criteria of five models for visualization and numerical comparison.

\subsection{Adaptive learning strategy for PDEs residuals}
\subsubsection{Fixed-Budget Online Adaptive Learning}
\begin{figure}[H]
    \centering
    \subfloat[After 5,000 iterations]{{\includegraphics[width=5.5cm]{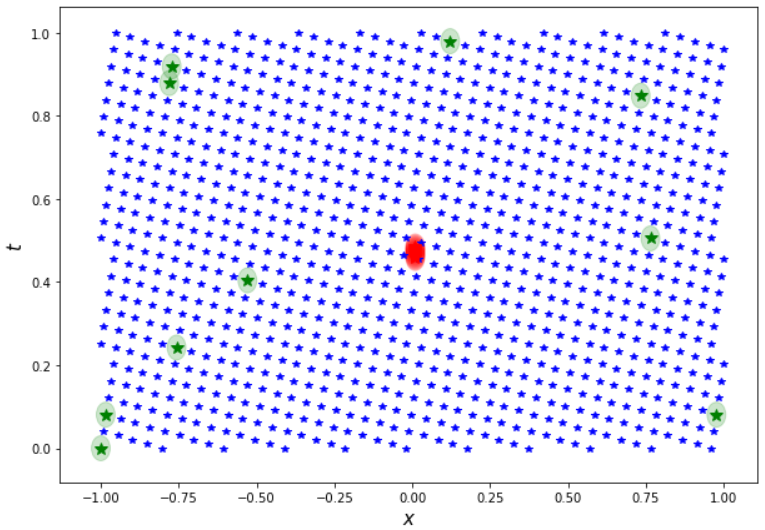} \label{slide_state1_prim}}}
    \subfloat[After 10,000 iterations]{{\includegraphics[width=5.5cm]{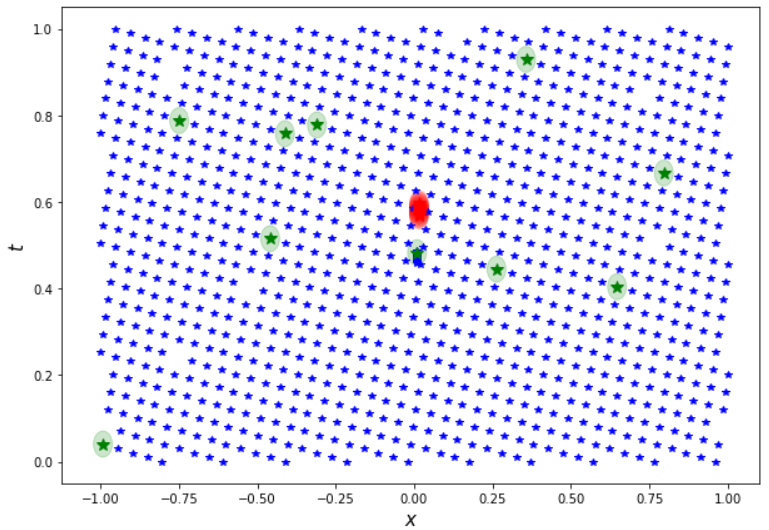} \label{slide_state2_prim}}}
    \subfloat[After 15,000 iterations]{{\includegraphics[width=5.5cm]{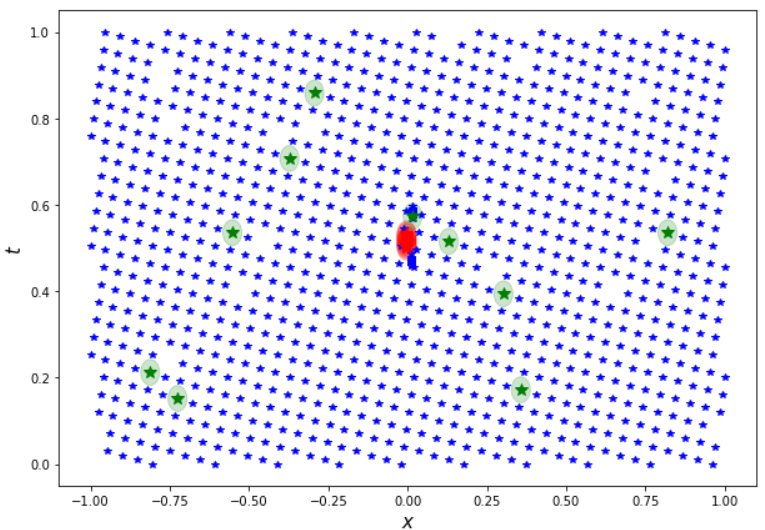} \label{slide_state3_prim}}}
    \caption{\textit{Illustration on Burgers equation of our primary idea: after every 5,000 iterations, we add 10 points (red points) and remove 10 points (green points) that yield the largest and the smallest PDE residuals. The blue points are the training collocations points of the previous 5,000 iterations.}}%
    \label{illus_primary_boal}%
\end{figure}

Motivated by the work of \cite{lu2021deepxde} which proposed 
the Residual-based Adaptive Refinement (RAR) that adds progressively during the training more collocation points at the locations that yield the largest PDE residuals, our primary idea is to control the number of training points potentially added by the method. We propose to remove the collocation points that yield the smallest PDE residuals so that the number of collocation points remains the same during the training. With this approach, the added points tend to be placed at nearly the same location corresponding to the global maximum value of the PDE residual (see Figure \ref{illus_primary_boal}). We propose to consider a set of sub-domains in order to capture not only the global extrema but also the local extrema of the PDEs residuals.


More precisely, we propose a Fixed-Budget Online Adaptive Learning (FBOAL) that adds and removes collocation points that yield the largest and the smallest PDE residuals on the sub-domains during the training. Algorithm \ref{fboam_algo} shows the details of this strategy. In FBOAL, the number of collocation points remains the same during the training, while their positions are relocated to the locations where the PDE residuals are important in each sub-domain. With the domain decomposition step, the algorithm is capable of detecting the local extrema inside the domain. Another remarkable advancement of this method is that in the parameterized problem, the collocation points can be relocated to the values of the parameter for which the solution is more complex (see section \ref{burgers_params} for an illustration on Burgers equation). To minimize the cost function, we adopt Adam optimizer with a learning rate decay strategy, which is proven to be very efficient in training deep learning models \citep{you2019does}, to increase the robustness of the model. More precisely, we observe that a high learning rate helps to faster the training but leads to an unstable cost function, while with a smaller learning rate, the model takes longer to train and has a stable cost function. The idea of the learning rate decay strategy is to begin the training with a high learning rate and decrease this value after a given number of iterations. In this work, we choose a set of learning rate values $lr=\{10^{-4}, 10^{-5}, 10^{-6}\}$. 

\begin{figure}[H]
    \centering
    \subfloat[Step 1]{{\includegraphics[width=5.5cm]{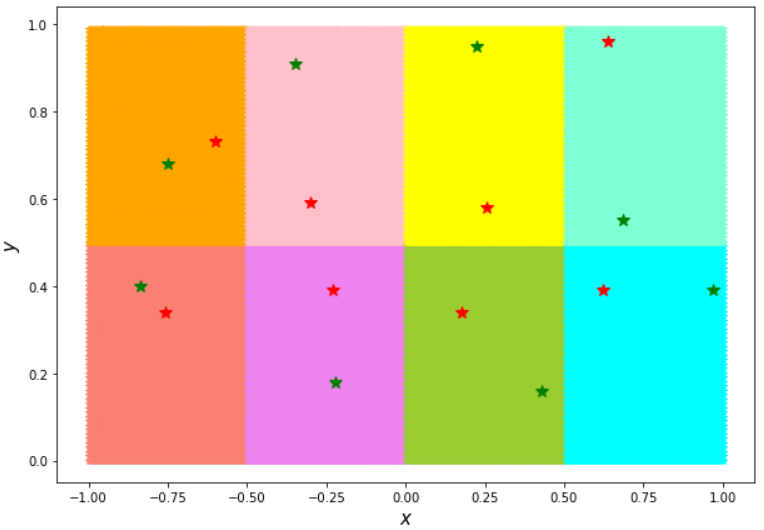} \label{slide_state1}}}
    \subfloat[Step 2]{{\includegraphics[width=5.5cm]{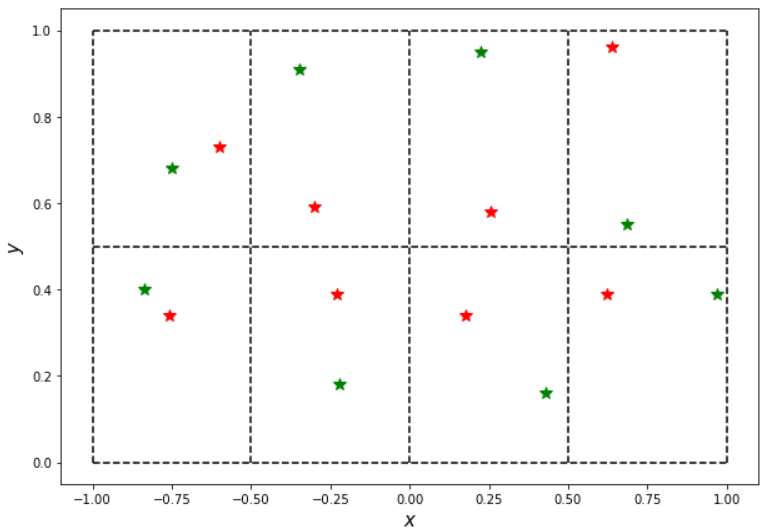} \label{slide_state2}}}
    \subfloat[Step 3]{{\includegraphics[width=5.5cm]{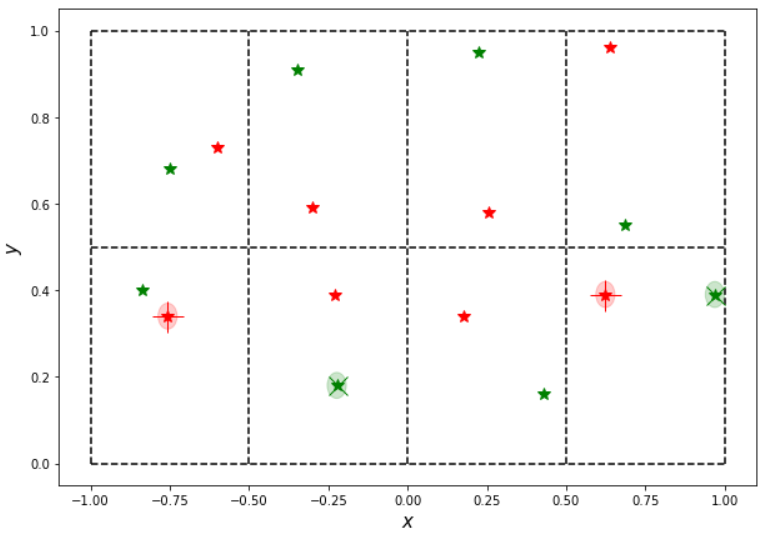} \label{slide_state3}}}
    \caption{\textit{FBOAL: Illustration of the algorithm on a rectangular domain $[-1,1]\times [0,1]$. In the first step (a), the domain is divided into smaller sub-domains (squares in this case). In each sub-domain, we calculate the PDE residuals and select the point which yields the largest residual (red points) and the point which yields the smallest residual (green points). In the second step (b), we assemble all the red points into the set $\mathcal{A}$ as described in the Algorithm \ref{fboam_algo} and all the green points into the set $\mathcal{R}$ as described in the Algorithm \ref{fboam_algo}. In the last step (c), we add $m=2$ points which yield the largest PDE residuals in the set $\mathcal{A}$ ("+" points in red) to the original set of collocation points $\mathcal{C}$, and remove $m=2$ points which yield the smallest PDE residuals in the set $\mathcal{R}$ ("x" points in green) from the set $\mathcal{C}$.  }}%
    \label{slide_state}%
\end{figure}

\begin{algorithm}[H]
\begin{algorithmic}[1]
\caption{Fixed-Budget Online Adaptive Learning (FBOAL)}
\label{fboam_algo}
\REQUIRE{The number of sub-domains $d$, the number of added and removed points $m$, the period of resampling $k$, a testing data set, a threshold $s$.}
\STATE Generate the set $\mathcal{C}$ of collocation points on the studied domain $\Omega$.
\STATE Divide the domain into $d$ sub-domains $\Omega_1 \cup \Omega_2 ... \cup \Omega_d = \Omega$.
\FOR{$lr_i$ in $lr$}
\REPEAT
\STATE Train PINNs for $k$ iterations with the learning rate $lr_i$.
\STATE Generate a new set $\mathcal{C'}$ of random points inside the domain $\Omega$.
\STATE Compute the PDE residual at all points in the set $\mathcal{C'}$ and the set $\mathcal{C}$.
\STATE On each subdomain $\Omega_i$, take 1 point of the set $\mathcal{C'}$ which yield the largest PDE residuals on the subdomain. Gather these points into a set $\mathcal{A}$.
\STATE On each subdomain $\Omega_i$, take 1 point of the set $\mathcal{C}$ which yield the smallest PDE residuals on the subdomain. Gather these points into a set $\mathcal{R}$.
\STATE Add $m$ points of the set $\mathcal{A}$ which yield the largest errors for the residuals to the set $\mathcal{C}$, and remove $m$ points of the set $\mathcal{R}$ which yield the smallest errors for the residuals.
\UNTIL The maximum number of iterations $K$ is reached or the error  of the prediction on the testing data set of reference is smaller than some threshold $s$.
\ENDFOR
\end{algorithmic}
\end{algorithm}
Figure \ref{slide_state} shows a simple illustration of the algorithm on a rectangular domain. We note that our primary idea which does not decompose the domain is a special case of the Algorithm \ref{fboam_algo} where each point of the initial collocation data set is added to the set $\mathcal{A}$ and $\mathcal{R}$. The way to divide the domain and the number of sub-domains can play important roles in the algorithm. If we dispose of expert knowledge on the PDEs problem, we can divide the domain as a finite-element mesh such that it is very fine at high-gradient locations and coarse elsewhere. In this primary work, we dispose of no knowledge \textit{a priori}.

\textbf{Discussion of the hyperparameters:}
\begin{itemize}
    \item The number of sub-domains $d$: This number has a great impact on the training time. A big value of $d$ leads to a large amount of time for the resampling. However, a small value of $d$ may decrease significantly the accuracy of the prediction if the local extrema of the PDE residuals are localized.
    \item The number of added and removed points $m$: This number can not be bigger than the number of sub-domains $d$. A high number of $m$ may decrease the effectiveness of the method as the algorithm may add unnecessary points and remove important points during the training. 
    \item The period of resampling $k$: this number has a great impact on the accuracy of the prediction. If this number is too small, the model may be not well-trained enough to accurately compute the PDEs residuals and this leads to the incorrect in adding and removing training points.
\end{itemize}

We will analyze in detail the impact of these hyperparameters in the next section. It is also worth noting that for the effectiveness of the algorithm, the number of points in the new set $\mathcal{C'}$ has to be much greater than in the set $\mathcal{C}$. This is because, with a smaller number of points in $\mathcal{C}'$, there is a high possibility that the largest errors in the set $\mathcal{C}'$ are smaller than the smallest errors in the set $\mathcal{C}$. To reduce this risk, in this work, we generate the set $\mathcal{C'}$ so that $|\mathcal{C'}|=10|\mathcal{C}|$, where $|.|$ defines the cardinal of the set. The stopping criterion is chosen based on a number of maximum iterations for each value of the learning rate and an error criterion computed on a testing data set of reference. The detail of this stopping criterion is specified in each use case. With this stopping criterion, the number of training iterations is also an adaptive number in each specific case, which helps to avoid unnecessary training iterations.

When dealing with multi-physics problems which involve systems of PDEs, we separate the set of collocation points into different subsets for each equation and then effectuate the process independently for each subset. Separating the set of collocation points helps to avoid the case when added points for one equation are removed for another.

\subsubsection{Residual-based Adaptive Distribution and Residual-based Adaptive Refinement with Distribution}
\cite{wu2023comprehensive} proposed two residual-based adaptive sampling approaches named Residual-based Adaptive Distribution (RAD) and Residual-based Adaptive Refinement with Distribution (RAR-D). Here, we briefly present the formulation of RAD and RAR-D which closely follows the framework introduced by \cite{wu2023comprehensive}. In these approaches, the probability density function (PDF) $p(x)$, which is based on the PDE residuals, is defined as follows:
\begin{align}
    p(x) \propto \dfrac{\epsilon^k(x)}{\mathbb{E}[\epsilon^k(x)]} + c \label{pdf_wu}
\end{align}
where $\epsilon(x)$ denotes the PDE residual at the points $x$, $k\ge 0$ and $c\ge 0$ are two hyperparameters.
\begin{itemize}
    \item Residual-based Adaptive Distribution (RAD): after a certain number of iterations, the collocation points are randomly resampled according to the PDF defined in equation (\ref{pdf_wu})
    \item Residual-based Adaptive Refinement with Distribution (RAR-D): after a certain number of iterations, $m$ points are randomly sampled according to the PDF defined in equation (\ref{pdf_wu}). Then these points are added to the set of training collocation points. 
\end{itemize}
\cite{wu2023comprehensive} also showed in detail the impact of the hyperparameters $k$ and $c$ on the performance of RAD and RAR-D in different PDEs setup. In this work, we do not illustrate in detail these impacts and only consider the values of $k$ and $c$ which yield the best accuracy of PINNs prediction. Besides that, it is worth noting that the performance of RAD and RAR-D also depends on the period of resampling. We do not investigate this impact and choose the same value of this number for these two approaches as for FBOAL for comparison.


The main differences between RAD, RAR-D, and our proposed method FBOAL lie in the domain decomposition step in FBOAL and the percentage of modified collocation points after every time we effectuate the re-sampling step. In fact, in FBOAL if we do not divide the domain into sub-domains, the goal of these methods becomes the same, i.e. all methodologies aim to assemble more collocation points at the position of the global maximum of the PDE residuals. However, while RAD suggests re-sampling all the training collocation points, FBOAL only modifies a certain percentage of points (which is controlled by the parameter $m$) during the process. Furthermore, the decomposition step helps FBOAL to be able to detect local maxima and local minima of the residuals inside the domain (which, however, depends on the way we divide the domain), and thus FBOAL gives equal concentration for all local maxima of the PDE residuals. 

\section{Numerical results}
In this section, we first compare and demonstrate the use of adaptive sampling methods (FBOAL, RAD, and RAR-D) to solve the Burgers equation in both non-parameterized context and parameterized context (i.e. the viscosity is fixed or not). We then illustrate the performance of FBOAL in a realistic industrial case: a system of PDEs that is used in the rubber calendering process. The results on a wave equation can be found in Appendix \ref{sec_wave}. The code in this study is available from the GitHub repository \url{https://github.com/nguyenkhoa0209/PINNs_FBOAL} upon publication.

In the following, unless specifying otherwise, we use the relative $\mathcal{L}^2$ error defined as below to compare the numerical performance of each methodology:
\begin{align*}
    \epsilon_w = \dfrac{||w-\hat{w}||_2}{||w||_2}
\end{align*}
where $w$ denotes the reference simulated field of interest and $\hat{w}$ is the corresponding PINNs prediction. 

\subsection{Burgers equation}
\begin{figure}[H]
    \centering
    \subfloat[For $\nu=0.0025$]{{\includegraphics[width=5.5cm]{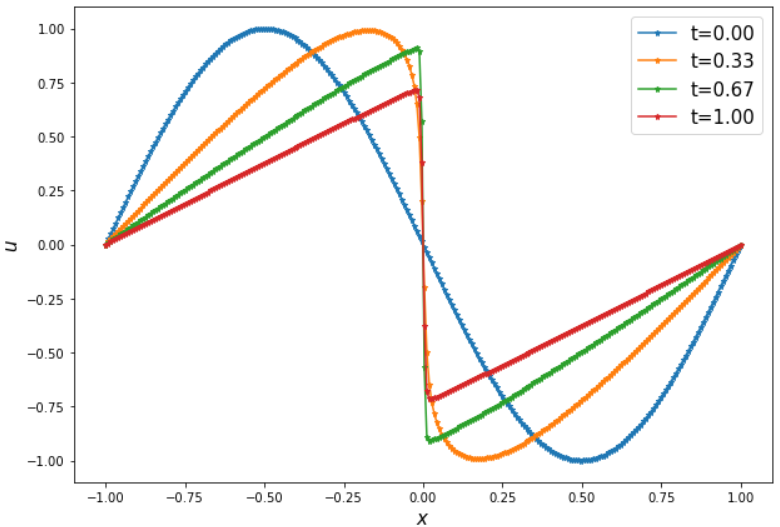} \label{burgers_fix_sol_nu0}}}
    \subfloat[For $\nu=0.0076$]{{\includegraphics[width=5.5cm]{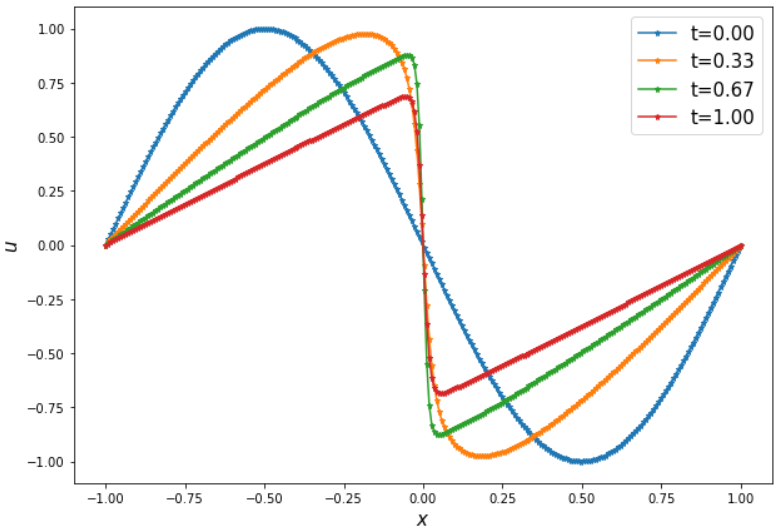} \label{burgers_fix_sol_nu1}}}
    \subfloat[For $\nu=0.0116$]{{\includegraphics[width=5.5cm]{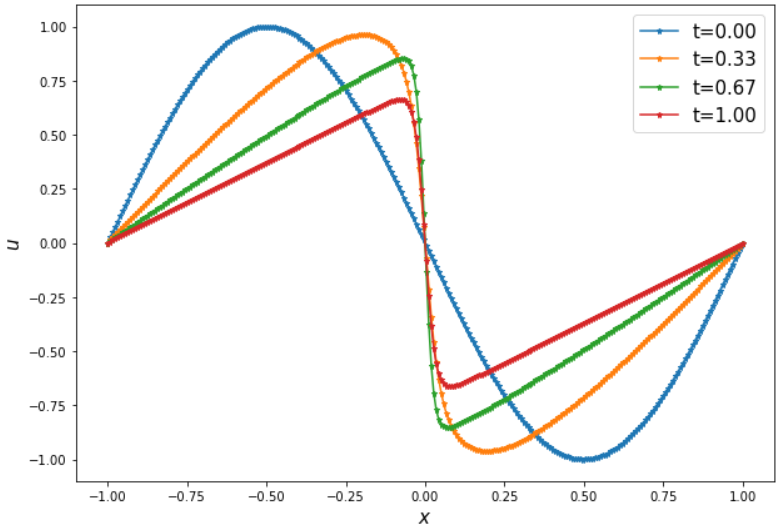} \label{burgers_fix_sol_nu2}}}
    \caption{\textit{Burgers equation: Reference solution at different instant $t$ for different values of $\nu$.}}%
    \label{burgers_sol}%
\end{figure}

We consider the following Burgers equation:
\begin{align*}
\begin{cases}
  u_t + uu_x -\nu u_{xx} = 0 \quad \text{for } x\in [-1,1], t\in[0, 1] \\
  u(x,0) = -\sin(\pi x)\\
  u(-1,t) = u(1,t) = 0
\end{cases}
\end{align*}
where $\nu$ is the viscosity. Figure \ref{burgers_sol} shows the reference solution for different values of $\nu$. For a small value of $\nu$, the solution is very steep close to $x=0$. For higher values, the solution becomes smoother. Thus when $\nu$ is fixed, it is expected that the collocation points are located close to $x=0$ during the training to better capture the discontinuity. When $\nu$ is varied, it is expected that the number of collocation points is more important for smaller values of $\nu$ while being located close to $x=0$. In the following, we assess whether FBOAL can relocate the collocation points to improve the performance of PINNs.

In our experiment, to simplify the cost function and guarantee the boundary and initial conditions, these conditions are forced to be automatically satisfied by using the following representation for the prediction $\bm{\hat{u}} =t(x-1)(x+1) \mathcal{NN}(.) - \sin(\pi x)$. The interested readers may refer to the work of \cite{berg2018unified, liu2019solving} for the general formulations. With this strategy, we do not need to adjust different terms in the loss function as there is only the loss for PDE residuals which is left. For the architecture of PINNs, we use a feedforward network with 4 hidden layers with 50 neurons per layer. To minimize the cost function, we adopt Adam optimizer with the learning rate decay strategy. The results are obtained with 50,000 epochs with the learning rate $lr = 10^{-3}$, 200,000 epochs with the learning rate $lr = 10^{-4}$ and 200,000 epochs with $lr = 10^{-5}$. For a fair comparison of all investigated methodologies, the initialization of the training collocation points is the same in all cases.

\subsubsection{Non-parameterized problem}
We first illustrate the performance of adaptive sampling approaches in a context where $\nu$ is fixed. We take $10$ equidistant values of $\nu\in[0.0025, 0.0124]$. For each $\nu$, we compare the performance of classical PINNs, PINNs with RAD, RAR-D, and PINNs with FBOAL. For the training of PINNs, we take $N_{pde}=1024$ collocation points that are initialized equidistantly inside the domain. We take a testing set of reference solutions on a $10\times 10$ equidistant spatio-temporal mesh and stop the training when either the number of iterations surpasses $K=500,000$ or the relative $\mathcal{L}^2$ error between PINNs prediction and the testing reference solution is smaller than the threshold $s=0.02$. The learning data set (which includes only the training collocation points) and the testing data set (which includes the testing points of reference) are independent. The following protocol allows us to compare fairly all adaptive sampling methods. For the training of FBOAL, we divide the domain into $d=200$ sub-domains as squares of size $0.1$. After every $k=2,000$ iterations, we add and remove $m= 2\%\times N_{pde}\approx 20$ collocation points based on the PDE residuals. For the training of RAD, we take $k=1$ and $c=1$ and effectuate the process after $2,000$ iterations. For the training of RAR-D, we take $k=2$ and $c=0$ and after every $2,000$ iterations, $5$ new points are added to the set of training collocation points. At the end of the training, the number of collocation points for FBOAL and RAD remains the same as at the beginning ($N_{pde}=1024$), while for RAR-D, this number increases gradually until the stopping criterion is satisfied.

\begin{figure}[H]
    \centering
    \subfloat[On the line $x=0$]{{\includegraphics[width=6.5cm]{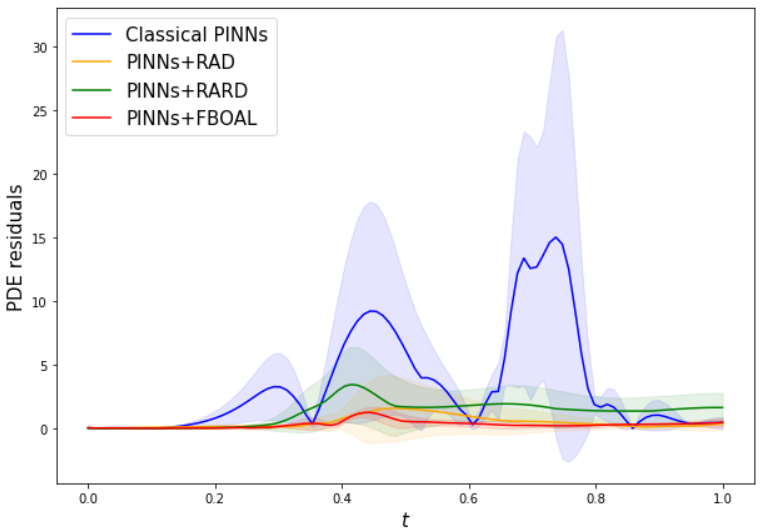} \label{burgers_fix_x0}}}
    \subfloat[At instant $t=1$]{{\includegraphics[width=6.5cm]{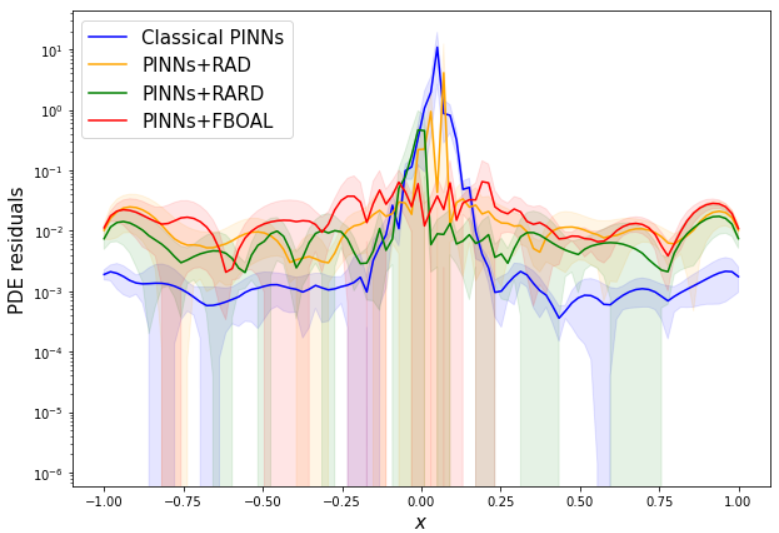} \label{burgers_fix_t1}}}
    \caption{\textit{Burgers equation: Absolute value of PDE residuals for $\nu=0.0025$ after the training process for different approaches.} The curves and shaded regions represent the geometric mean and one standard deviation of five runs.}%
    \label{burgers_fix_compare_pde}%
\end{figure}

Figure \ref{burgers_fix_compare_pde} shows the PDE residuals for $\nu=0.0025$ after the training process on the line $x=0$ and at the instant $t=1$ where the solution is very steep. On the line $x=0$ (Figure \ref{burgers_fix_x0}), with classical PINNs where the collocation points are fixed during the training, the PDE residuals are very high and obtain different peaks (local maxima) at different instants. It should be underlined that all the considered adaptive sampling methods are able to decrease the values of local maxima of the PDEs residuals. Among the approaches where the number of collocation points is fixed, FBOAL is able to obtain the smallest values for the PDE residuals, and for its local maxima, among all the adaptive resampling methodologies. At the instant $t=1$ (Figure \ref{burgers_fix_t1}), the same conclusion can be drawn.

\begin{figure}[H]
    \centering
    \subfloat[Relative $\mathcal{L}^2$ error]{{\includegraphics[width=5.5cm]{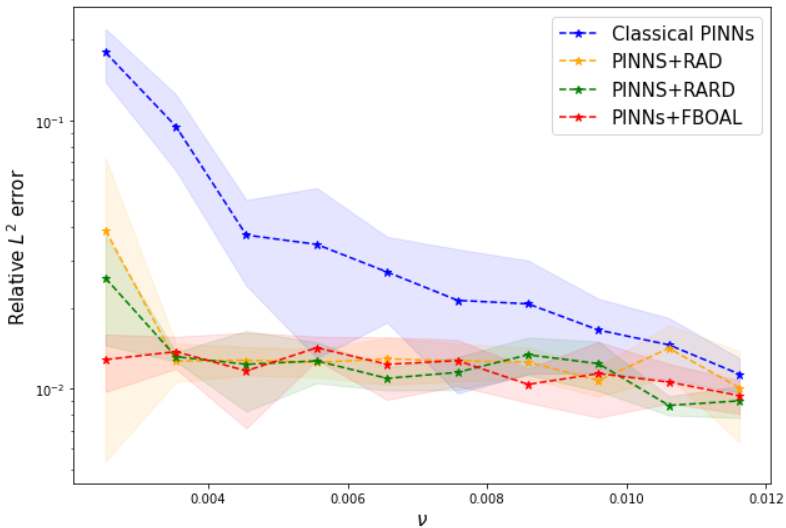} \label{burgers_fix_err}}}
    \subfloat[Number of training iterations]{{\includegraphics[width=5.7cm]{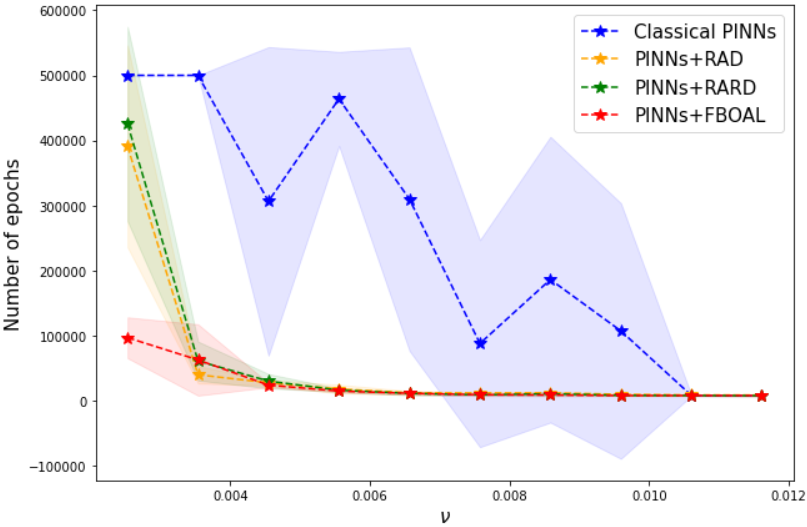} \label{burgers_fix_nb}}}
    \subfloat[Loss for $\nu=0.0025$]{{\includegraphics[width=5.5cm]{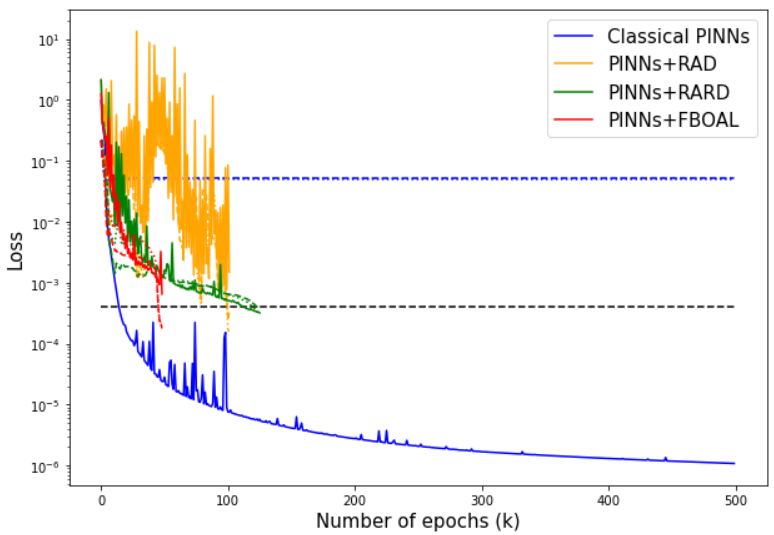} \label{burgers_fix_cost}}}
    \caption{\textit{Burgers equation: Comparison of classical PINNs and PINNs with adaptive sampling approaches.} The curves and shaded regions represent the geometric mean and one standard deviation of five runs. In (c) the solid lines show the cost function during the training, the dashed lines show the errors on the testing mesh, and the black line shows the threshold to stop the training.}%
    \label{burgers_fix_compare}%
\end{figure}

To assess the overall performance of PINNs, we evaluate the errors of the prediction on a $256 \times 100$ spatio-temporal mesh (validation mesh). Figure \ref{burgers_fix_err} shows the relative $\mathcal{L}^2$ error between PINNs predictions and reference solution. Figure \ref{burgers_fix_nb} shows the number of training iterations for PINNs to meet the stopping criterion. As expected, when $\nu$ increases, which means the solution becomes smoother, the accuracy of PINNs in all methodologies increases and the models need less number of epochs to meet the stopping criterion. We note that when $\nu$ is large and the solution is very smooth, the classical PINNs are able to give comparable performance to PINNs with adaptative methodologies. However, when $\nu$ becomes smaller, it is clear that PINNs with adaptive sampling methods outperform classical PINNs in terms of accuracy and robustness. Among these strategies, FBOAL provides the best accuracy in terms of errors and also needs the least iterations to stop the algorithm. Table \ref{time_fix} illustrates the training time of each methodology for $\nu=0.0025$. We observe that by using FBOAL a huge amount of training time is gained compared to other approaches. Figure \ref{burgers_fix_cost} illustrates the cost function during the training process for $\nu=0.0025$ and the errors of the prediction on the testing mesh. For clarity, only the best cost function (which yields the smallest values after the training process in five runs) of each methodology is plotted. After every $k=2,000$ epochs, as the adaptive sampling methods relocate the collocation points, there are jumps in the cost function. The classical PINNs minimize the cost function better than other methodologies because the position of collocation points is fixed during the training. This leads to the over-fitting of classical PINNs on the training collocation points and does not help to increase the accuracy of the prediction on the testing mesh. While with adaptive sampling methods, the algorithm achieves better performance on the generalization to different meshes (the testing and validation meshes).

Table \ref{time_fix} provides the training time and the number of resampling of each methodology. We see that the resampling does affect the training time. More precisely, with RAR-D and RAD, a huge number of resampling is effectuated, which leads to a long training time compared to the classical PINNs (even though these methods need smaller numbers of training iterations). While with FBOAL, the number of resampling is small and we obtain a smaller training time compared to the classical PINNs.

\begin{table}[H]
\centering
\begin{tabular}{ |c|c|c|c|c| } 
\hline
 & Classical PINNs & PINNs + RAR-D & PINNs + RAD & PINNs + FBOAL\\
\hline
Training time &  33.7 $\pm$ 1.5 & 41.0 $\pm$ 5.8 & 38.8 $\pm$ 2.3 & \textbf{21.5} $\pm$ \textbf{2.7} \\
Number of resampling & 0 $\pm$ 0 & 201 $\pm$ 75 & 210 $\pm$ 77 & \textbf{48} $\pm$ \textbf{2} \\
\hline
\end{tabular}
\caption{\textit{Burgers equation: Training time (in minutes) and the number of resampling for $\nu=0.0025$.} The training is effectuated on an NVIDIA V100 GPU card.}
\label{time_fix}
\end{table}

\begin{figure}[H]
    \centering
    \subfloat[For $\nu=0.0025$]{{\includegraphics[width=5.5cm]{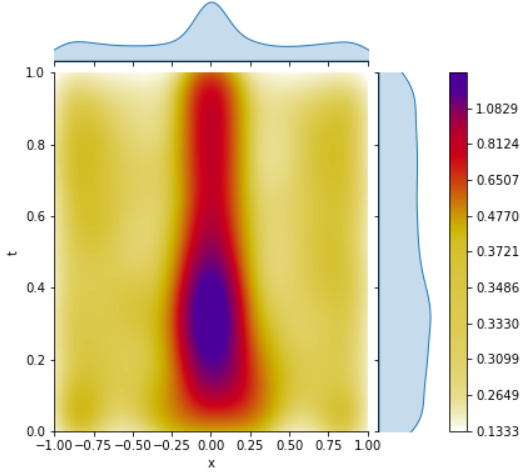} \label{burgers_fix_dnu0}}}
    \subfloat[For $\nu=0.0076$]{{\includegraphics[width=5.5cm]{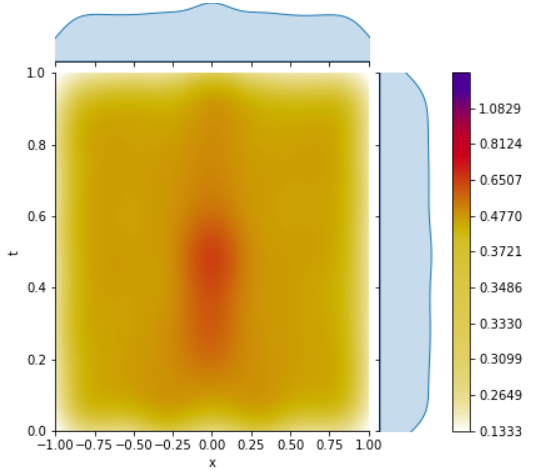} \label{burgers_fix_dnu1}}}
    \subfloat[For $\nu=0.0116$]{{\includegraphics[width=5.5cm]{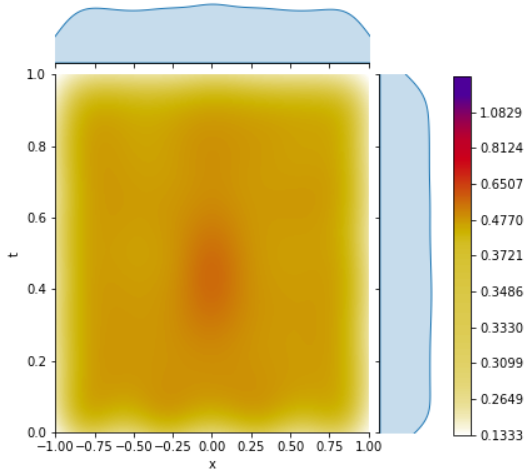} \label{burgers_fix_dnu2}}}
    \caption{\textit{Burgers equation: Density of collocation points after the training with FBOAL.}}%
    \label{burgers_fix_dens}%
\end{figure}

In the following, we analyze in detail the performance of FBOAL. Figure \ref{burgers_fix_dens} illustrates the density of collocation points after the training with FBOAL for different values of $\nu$. We see that, for the smallest value $\nu=0.0025$, FBOAL relocates the collocation points close to x=0 where the solution is highly steep. For $\nu=0.0076$, as there is only one iteration of FBOAL that adds and removes points (see Figure \ref{burgers_fix_nb}), there is not much difference with the initial collocation points but we still see few points are added to the center of the domain, where the solution becomes harder to learn. For the biggest value $\nu=0.0116$, there is no difference with the initial collocation points as PINNs already satisfy the stopping criterion after a few iterations.

\begin{figure}[H]
    \centering
    \subfloat[When $d=200$, $k=2,000$]{{\includegraphics[width=5.5cm]{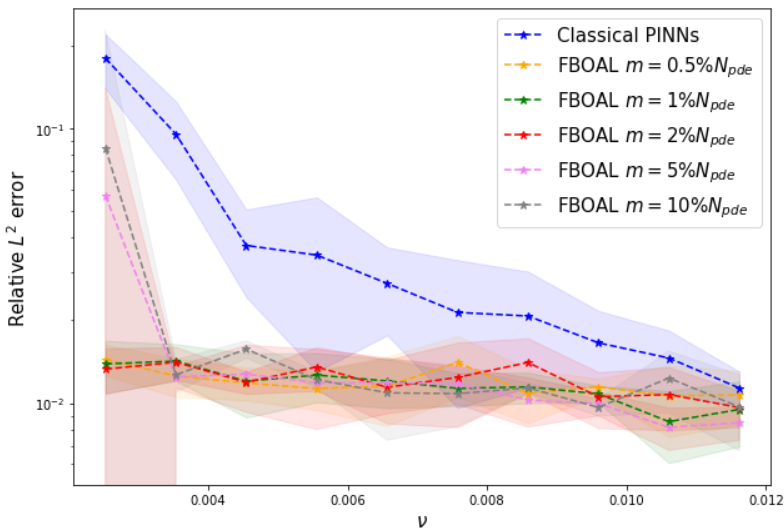} \label{burgers_fix_varyall_1}}}
    \subfloat[When $d=50$, $m=0.5\%N_{pde}$]{{\includegraphics[width=5.4cm]{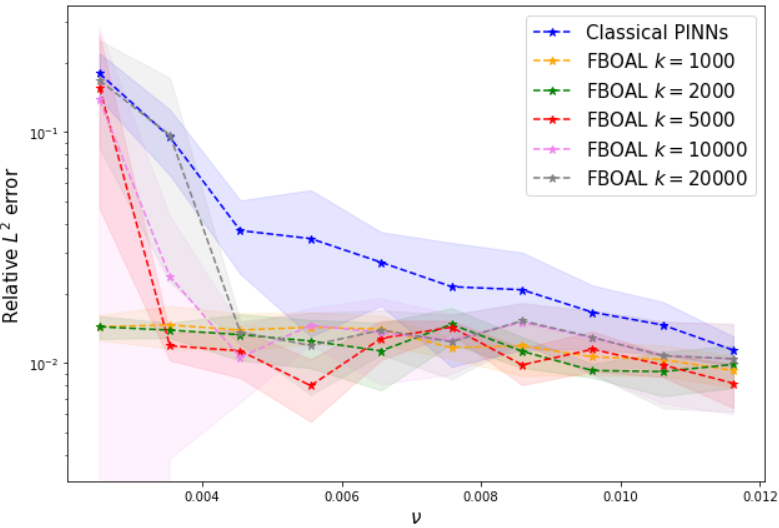} \label{burgers_fix_varyall_2}}}
    \subfloat[When $m=0.5\%N_{pde}$, $k=1,000$]{{\includegraphics[width=5.5cm]{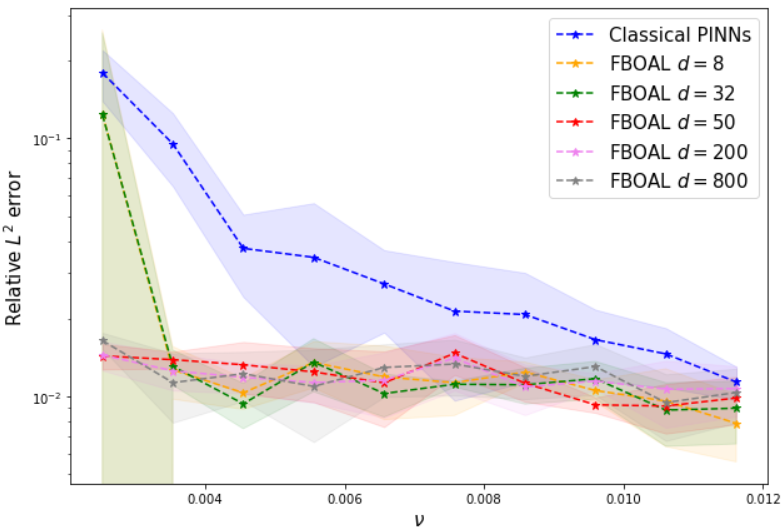} \label{burgers_fix_varyall_3}}}
    \quad
    \caption{\textit{Burgers equation: Performance in terms of accuracy of FBOAL when varying $m$, $k$, and $d$.} The curves and shaded regions represent the geometric mean and one standard deviation of five runs.}%
    \label{burgers_boal_varyall}%
\end{figure}

\begin{figure}[H]
    \centering
    \subfloat[When $d=200$, $k=2,000$]{{\includegraphics[width=5.5cm]{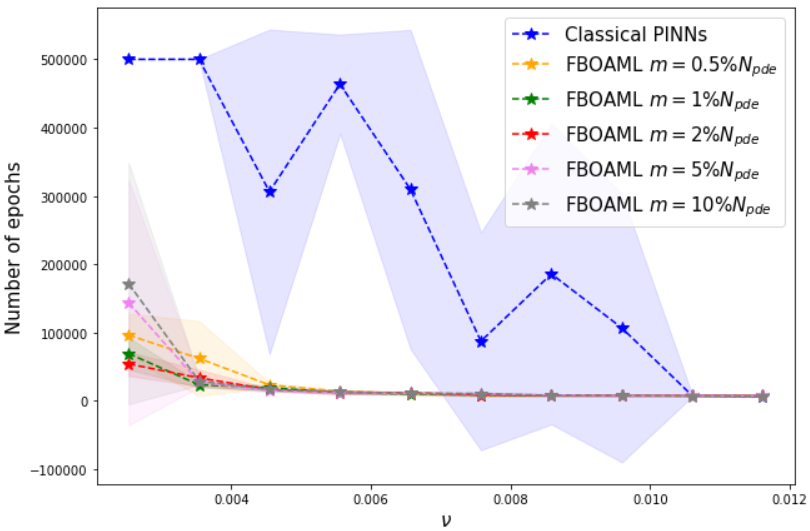} \label{burgers_fix_varyall_nb_1}}}
    \subfloat[When $d=50$, $m=0.5\%N_{pde}$]{{\includegraphics[width=5.4cm]{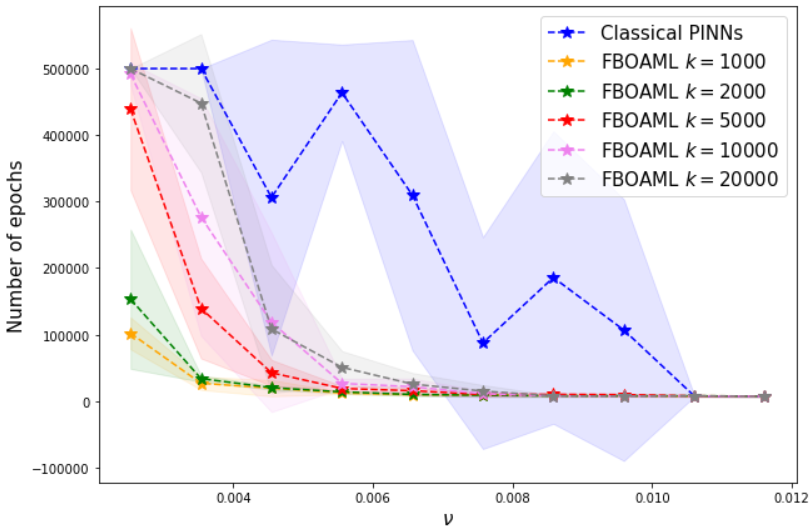} \label{burgers_fix_varyall_nb_2}}}
    \subfloat[When $m=0.5\%N_{pde}$, $k=1,000$]{{\includegraphics[width=5.5cm]{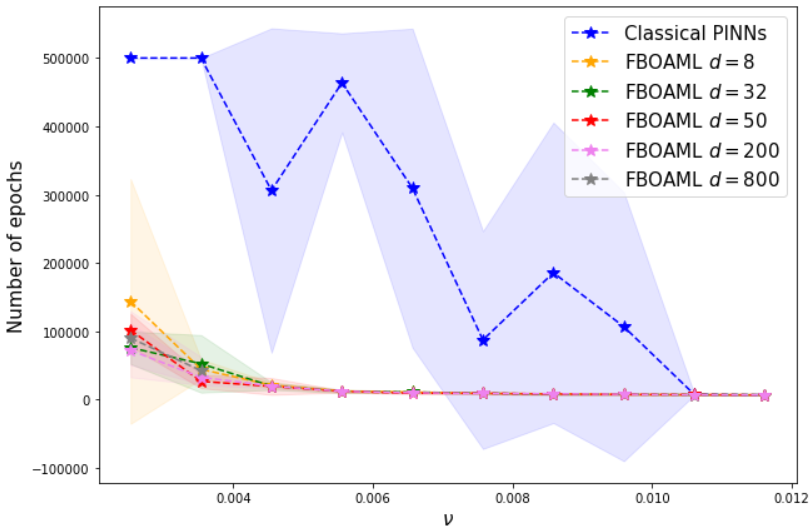} \label{burgers_fix_varyall_nb_3}}}
    \quad
    \caption{\textit{Burgers equation: Number of training iterations of FBOAL when varying $m$, $k$, and $d$.} The curves and shaded regions represent the geometric mean and one standard deviation of five runs.}%
    \label{burgers_boal_varyall_nb}%
\end{figure}

We next test the performance of FBOAL when varying $m$ (the number of added and removed points), $k$ (the period of resampling), and $d$ (the number of sub-domains). Figures \ref{burgers_boal_varyall} and \ref{burgers_boal_varyall_nb} illustrate the relative $\mathcal{L}^2$ errors and the number of training iterations produced by different PINNs combined with FBOAL when varying $m$, $k$ and $d$, respectively. More results with different values of these hyper-parameters are provided in Appendix \ref{annex_burgers_fix}. It can be seen from Figure \ref{burgers_fix_varyall_1} that a big value of $m$ can decrease significantly the accuracy of the prediction, and the model needs more training iterations to reach the stopping criteria. The same observation can be seen with a high value of $k$ (Figure \ref{burgers_fix_varyall_2}). While the impact of the parameter $d$ is hard to interpret as it strongly depends on the values of $k$ and $m$ ( Figures \ref{burgers_fix_varyall_3} and \ref{burgers_boal_varyd}). Furthermore, this number also has a great impact on the training time. More precisely, a big value of $d$ leads to a large amount of time for the resampling. However, with a small value of $k$, we see that the sensitivity of the performance of PINNs to different values of $d$ can be decreased. In conclusion, we suggest starting with small values of $m$ and $k$ and then increasing these values later to see whether the predictions can become more accurate or not.


\subsubsection{Parameterized problem}\label{burgers_params}
We now illustrate the performance of PINNs in a parameterized problem where $\nu$ can be varied. In this case, $\nu$ is also considered as an input of PINNs, i.e. the network in PINNs is represented as $\mathcal{NN}(\bm{\mathrm{x}}, t, \nu, \bm{\theta})$. With this configuration, PINNs can predict the solution for different values of $\nu$ in one model. For the training, we take $40$ values of $\nu\in[0.0025,0.0124]$. For each $\nu$, we initialize 1024 equidistant collocation points, which leads to $N_{pde}=1024 \times 40= 40,960$ collocation points in total. During the training with FBOAL, the number of total collocation points remains the same, however, the number of collocation points can be varied for each $\nu$. We take a testing set of reference solutions on a $10\times 10$ equidistant spatio-temporal mesh and stop the training when either the number of iterations surpasses $K=2 \times 10^6$ or the sum of relative $\mathcal{L}^2$ error between PINNs prediction and the testing reference solution of all learning values of $\nu$ is smaller than the threshold $s=0.02\times 40=0.8$. We note again that the learning data set (which includes only the collocation points) and the testing data set (which includes the testing points of reference) are independent. For the training of FBOAL, we divide the domain into $d=200$ sub-domains as squares of size $0.1$. After every $k=2,000$ iterations, we add and remove $m=0.5\%\times N_{pde}\approx 200$ collocation points based on the PDE residuals. For the training of RAD, we take $k=1$ and $c=1$ and effectuate the process after $2,000$ iterations. For the training of RAR-D, we take $k=2$ and $c=0$ and after every $2,000$ iterations, $5\times 40=200$ new points are added to the set of training collocation points. At the end of the training, the number of collocation points for FBOAL and RAD remains the same as the beginning ($N_{pde}=40,960$), while for RAR-D, this number increases gradually until the stopping criterion is satisfied.

\begin{figure}[H]
    \centering
    \subfloat[On the line $x=0$]{{\includegraphics[width=6.5cm]{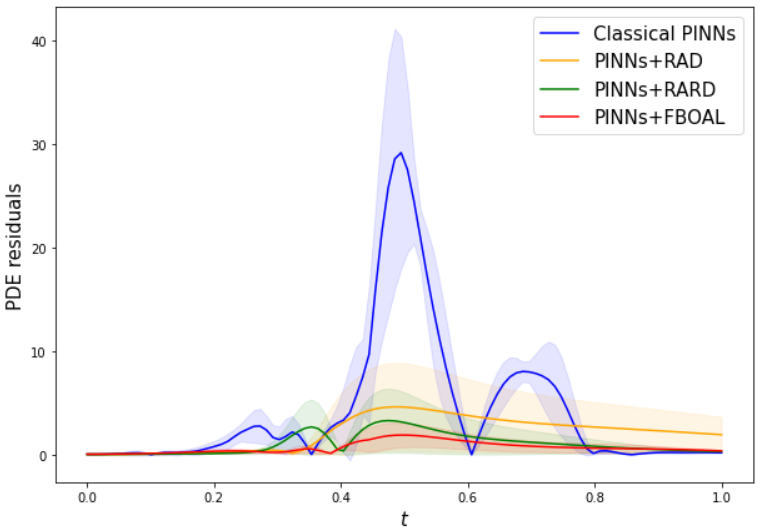} \label{burgers_params_x0}}}
    \subfloat[At instant $t=1$]{{\includegraphics[width=6.5cm]{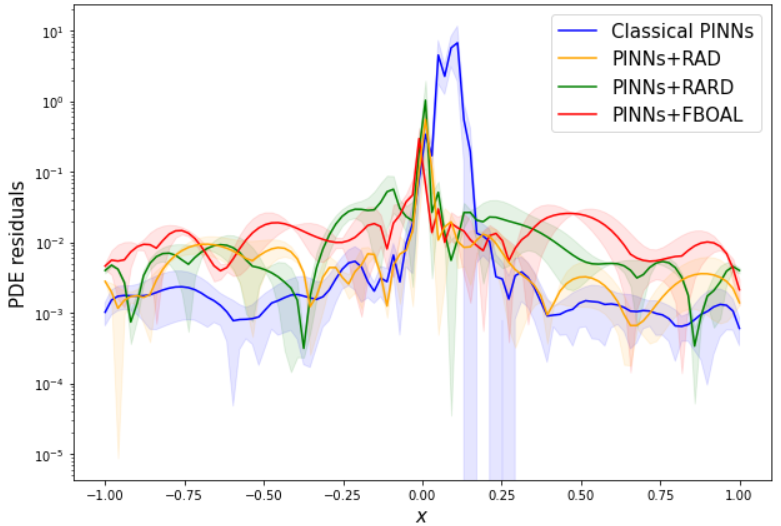} \label{burgers_params_t1}}}
    \caption{\textit{Burgers equation: Absolute value of PDE residuals for $\nu=0.0025$ after the training process by different approaches.} The curves and shaded regions represent the geometric mean and one standard deviation of five runs.}%
    \label{burgers_params_compare_pde}%
\end{figure}

\begin{figure}[H]
    \centering
    \subfloat[Relative $\mathcal{L}^2$ error]{{\includegraphics[width=5.5cm]{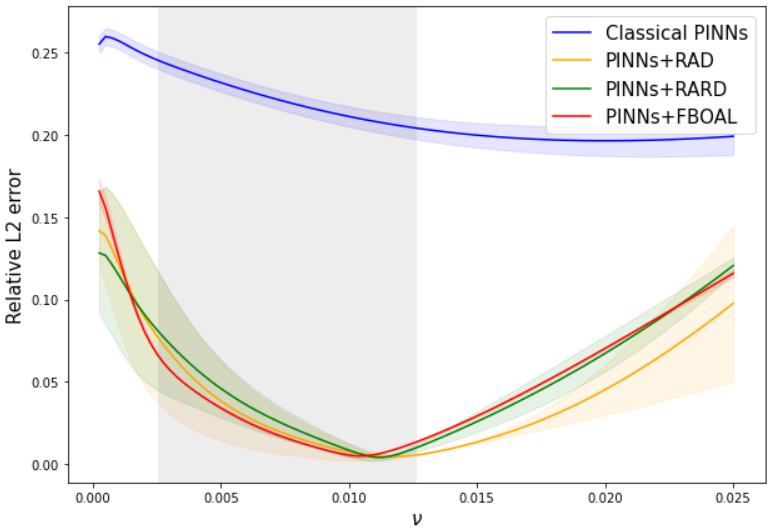} \label{burgers_params_err}}}
    \subfloat[Number of collocation points]{{\includegraphics[width=5.5cm]{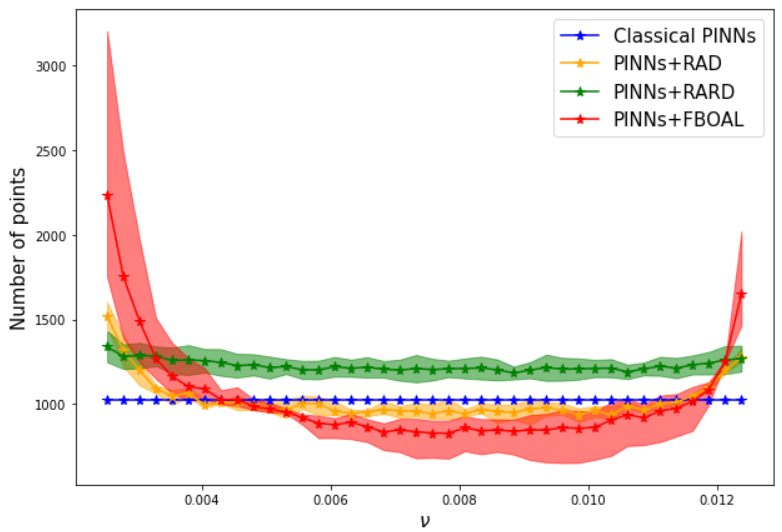} \label{burgers_params_nb}}}
    \subfloat[Cost function during the training]{{\includegraphics[width=5.5cm]{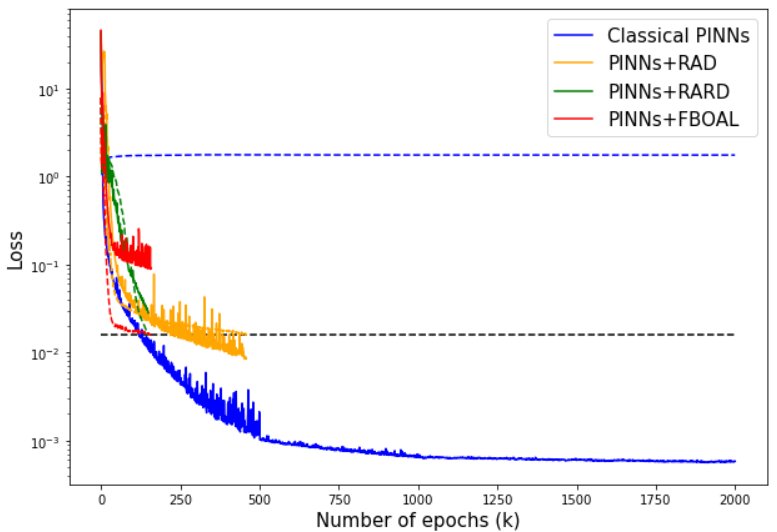} \label{burgers_params_cost}}}
    \caption{\textit{Burgers equation: comparison of classical PINNs and PINNs with adaptive sampling approaches.} The zone in gray is the learning interval for $\nu$ (interpolation zone). The curves and shaded regions represent the geometric mean and one standard deviation of five runs. In (c) the solid lines show the cost function during the training, the dashed lines show the errors on the testing data set, and the black line shows the threshold to stop the training.}%
    \label{burgers_params_compare}%
\end{figure}

Figure \ref{burgers_params_compare_pde} illustrates the absolute value of the PDE residuals for $\nu=0.0025$ on the line $x=0$ and at the instant $t=1$ where the solution is very steep. On the line $x=0$ (Figure \ref{burgers_params_x0}), with classical PINNs, the PDE residuals are very high and obtain different local maxima at different instants. Again, all the considered adaptive sampling methods are able to decrease the values of local maxima of the PDEs residuals. Among these approaches, FBOAL is able to obtain the smallest values for the PDE residuals. We note that the number of collocation points of each methodology is different (see Figure \ref{burgers_params_nb}). For $\nu=0.0025$, FBOAL relocates a very high number of collocation points. This leads to a better performance of FBOAL for $\nu=0.0025$. At the instant $t=1$, FBOAL, RAD and RAR-D provide nearly the same performance and are able to decrease the values of local maxima of the PDEs residuals compared to the classical PINNs.

Figure \ref{burgers_params_err} shows the relative $\mathcal{L}^2$ error between PINNs prediction and reference solution on a $256 \times 100$ mesh. The zone in gray represents the learning interval for $\nu$ (interpolation zone). Figure \ref{burgers_params_nb} shows the number of collocation points for each $\nu$. We see again in the interpolation zone, that when $\nu$ increases, the accuracy of PINNs in all methodologies increases. It is clear that PINNs with adaptive sampling methods outperform classical PINNs in terms of accuracy for all values of $\nu$ in both interpolation and extrapolation zones. As expected, with the approaches FBOAL and RAD (where $N_{pde}$ is fixed), the number of collocation points for smaller values of $\nu$ is more important than the number for higher ones. This demonstrates the capability of FBOAL and RAD of removing unnecessary collocation points for higher values of $\nu$ (whose solution is easier to learn) and adding them for smaller values of $\nu$ (as the solution becomes harder to learn). When $\nu$ tends to the higher extreme, the number of training points increases as there are fewer training values for $\nu$ in this interval. While with RAR-D (where $N_{pde}$ increases gradually), we do not see the variation of the number of collocation points between different values of $\nu$ as RAR-D tries to add more points at the discontinuity for all $\nu$. Table \ref{time_param} provides the training time of and the number of resampling of each methodology. It is clear that RAR-D outperforms the other approaches in terms of computational time as this approach needs very few numbers of resampling to meet the stopping criterion. However, to achieve this gain, RAR-D added for about 40,000 new collocation points after the training, which is not profitable in terms of memory. RAD and FBOAL provide comparable computational training time in this case and they both outperform the classical PINNs while the number of collocation points in total is fixed. Figure \ref{burgers_params_cost} illustrates the cost function during the training process and the errors of the prediction on the testing mesh. For clarity, only the best cost function (which yields the smallest number of iterations after the training process in five runs) of each methodology is plotted. Again, since the position of the collocation points is fixed during the training, classical PINNs minimize the cost function better than PINNs with adaptive sampling methods, which leads to over-fitting on the training data set and does not help to improve the accuracy of the prediction on the testing mesh. We see that FBOAL and RAR-D need a much smaller number of iterations to meet the stopping criterion.

\begin{table}[H]
\centering
\begin{tabular}{ |c|c|c|c|c| } 
\hline
 & Classical PINNs & PINNs + RAR-D & PINNs + RAD & PINNs + FBOAL\\
\hline
Training time &  13.2 $\pm$ 0.0 & \textbf{1.4} $\pm$ \textbf{0.3} & 9.1 $\pm$ 3.5 & 7.8 $\pm$ 1.9 \\
Number of resampling & 0 $\pm$ 0 & \textbf{34} $\pm$ \textbf{8} & 204 $\pm$ 71 & 173 $\pm$ 25 \\
\hline
\end{tabular}
\caption{\textit{Burgers equation: Training time (in hours) and the number of resampling of each methodology.} The training is effectuated on an NVIDIA A100 GPU card.}
\label{time_param}
\end{table}

\begin{figure}[H]
    \centering
    \subfloat[For $\nu=0.0025$]{{\includegraphics[width=5.5cm]{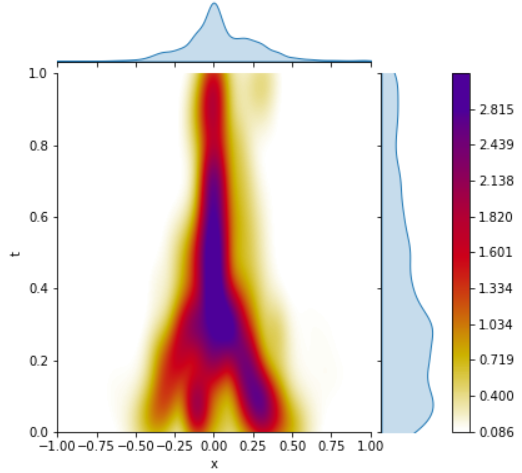} \label{burgers_params_dnu0}}}
    \subfloat[For $\nu=0.0076$]{{\includegraphics[width=5.5cm]{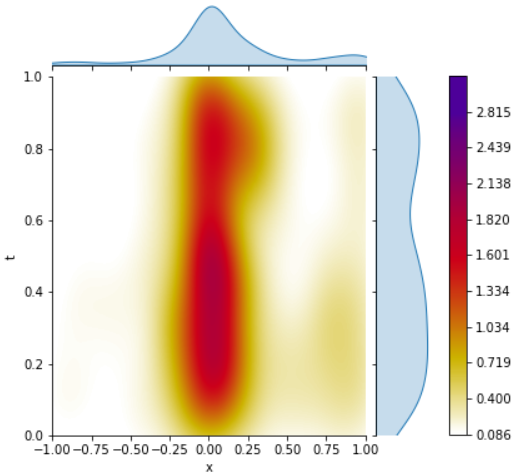} \label{burgers_params_dnu1}}}
    \subfloat[For $\nu=0.0116$]{{\includegraphics[width=5.5cm]{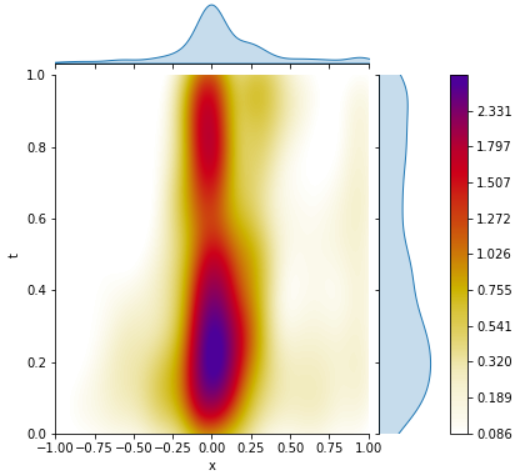} \label{burgers_params_dnu2}}}
    \caption{\textit{Burgers equation: Density of collocation points after the training by FBOAL.}}%
    \label{burgers_params_dens}%
\end{figure}

In the following, we analyze in detail the performance of FBOAL. Figure \ref{burgers_params_dens} illustrates the density of collocation points after the training with FBOAL with different values of $\nu$. We see that for all values of $\nu$, FBOAL relocates the collocation points close to the discontinuity.

\begin{figure}[H]
    \centering
    \subfloat[When $d=200$, $k=5,000$]{{\includegraphics[width=5.5cm]{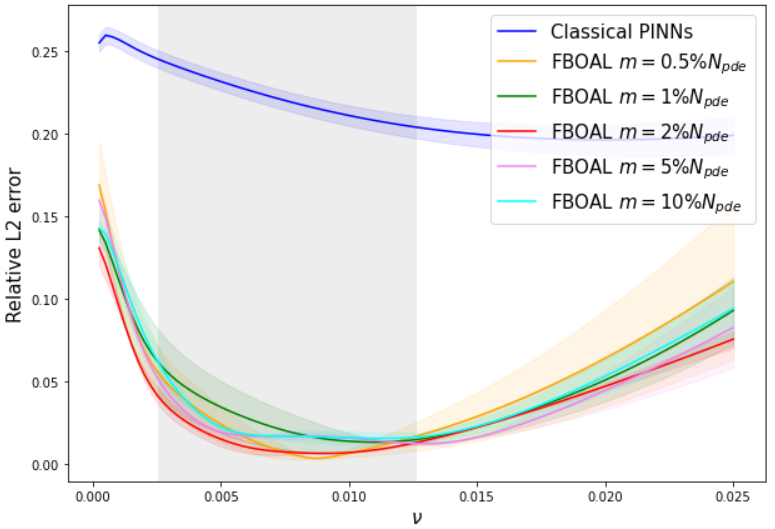} \label{burgers_params_varym}}}
    \subfloat[When $d=50$, $m=1\%N_{pde}$]{{\includegraphics[width=5.5cm]{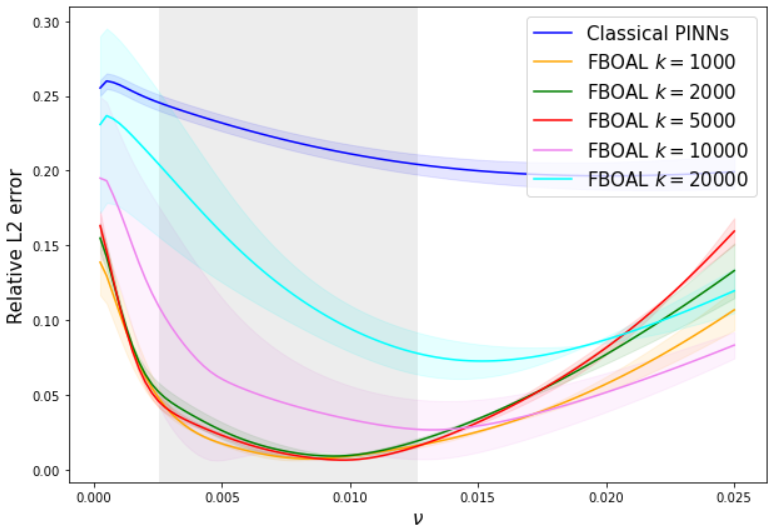} \label{burgers_params_varyk}}}
    \subfloat[When $k=2,000$, $m=0.5\%N_{pde}$]{{\includegraphics[width=5.5cm]{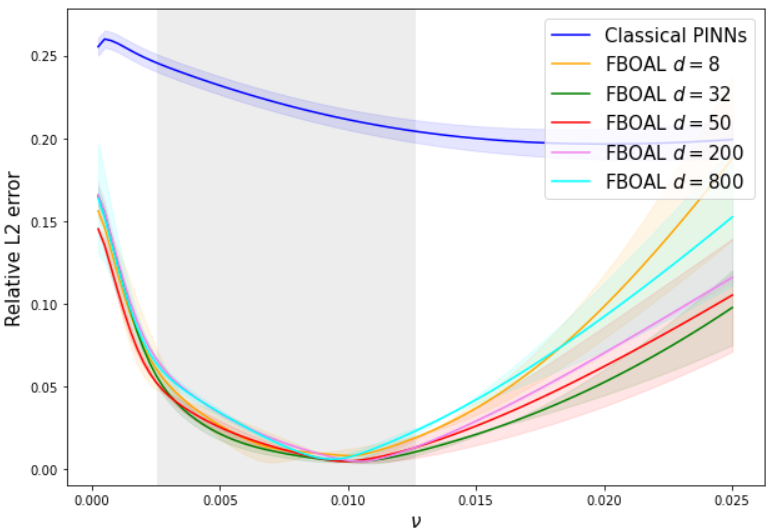} \label{burgers_params_varyd}}}
    \caption{\textit{Burgers equation: Performance in terms of accuracy of FBOAL when varying $m$, $k$ and $d$.} The zone in gray is the learning interval for $\nu$ (interpolation zone). The curves and shaded regions represent the geometric mean and one standard deviation of five runs.}%
    \label{burgers_boal_params_vary}%
\end{figure}

We test the performance of FBOAL when varying $m$, $k$, and $d$. Figure \ref{burgers_boal_params_vary} shows the relative $\mathcal{L}^2$ errors by different models when varying $m$, $k$, and $d$.  More results with different values of these hyper-parameters are provided in Appendix \ref{annex_burgers_param}. In general, with FBOAL, PINNs provide higher accuracy for the prediction than the classical PINNs with all considered values of $m$, $k$, and $d$ in both interpolation and extrapolation. However, it can be seen that these values still have important impacts on the performance of FBOAL. We can see that a high value of $k$ can affect negatively the accuracy. And with a small value of $k$, the performance of PINNs becomes less sensible to the values of $m$ and $d$. For the utilization of FBOAL, we suggest starting with small values of $m$ and $k$, and then increasing these values later to see whether the predictions can become more accurate or not.

\subsection{Application to calendering process}
\begin{figure}[H]
    \centering
    \includegraphics[width=10cm]{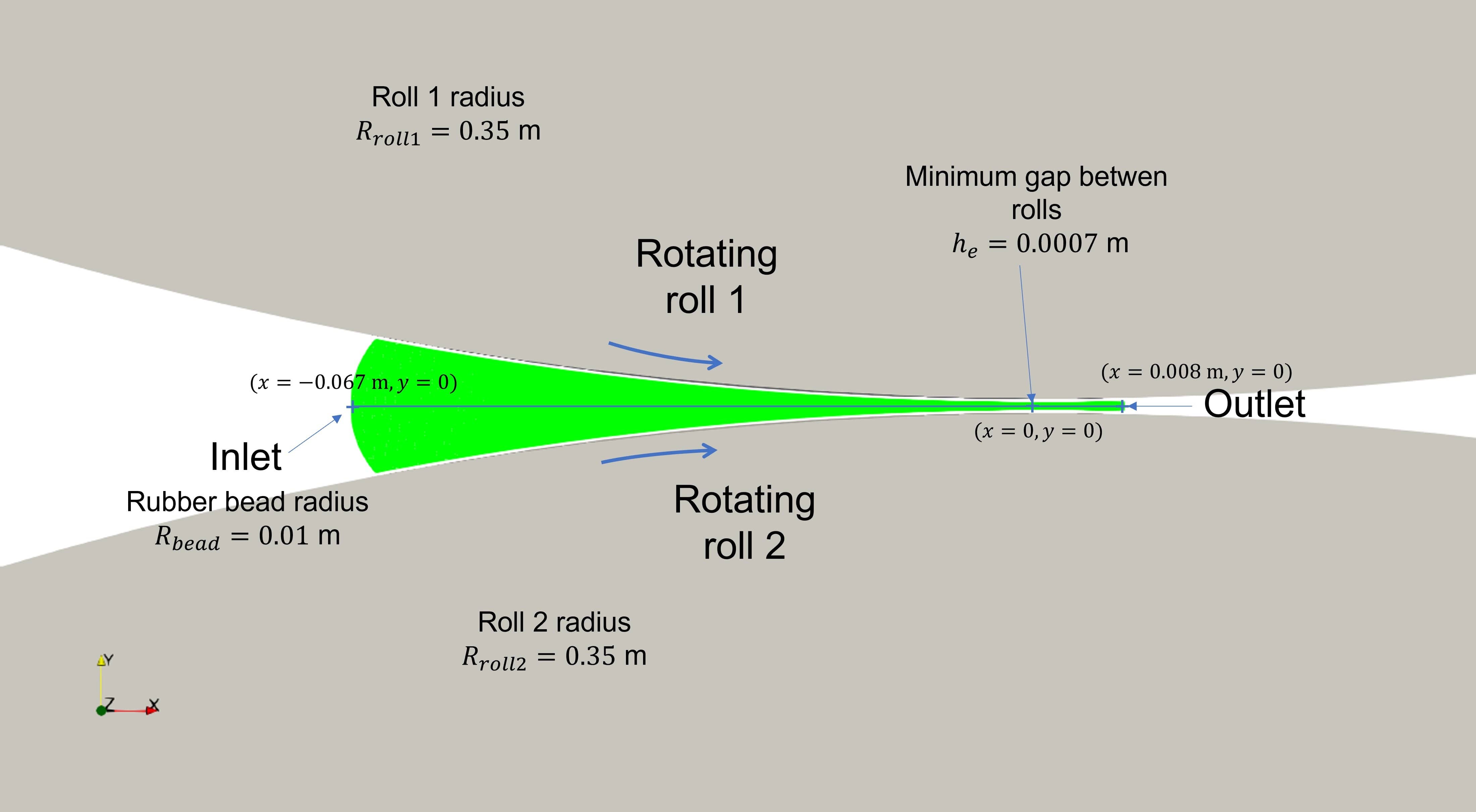}
    \caption{Sketch of the 2D configuration and geometrical setup considered in the present study. The physical domain of interest is colored in green.}%
    \label{calender2d}%
\end{figure}

In the industry of tire manufacturing, calendering is a mechanical process used to create and assemble thin tissues of rubber. From the physical point of view, the rubber is assimilated as an incompressible non-Newtonian fluid flow in the present study (in particular, the elastic part of the material is not considered here). Moreover, only the steady-state regime is considered. The goal of the present study is only to model the 2D temperature, velocity, and pressure fields inside rubber materials going through two contra-rotating rolls of the calender as depicted in Figure \ref{calender2d}. 

A detailed description of the calendering process and its mathematical formulation can be found in \cite{nguyen2022physics}. Here we briefly introduce the dimensionless PDEs system that is used in PINNs training process:
\begin{align*}
     &\tilde{\nabla}.\Big(2\tilde{\eta}(\bm{\vec{\tilde{u}}}, \tilde{T})\tilde{\bar{\bar{\epsilon}}}(\bm{\vec{\tilde{u}}})\Big) - \vec{\tilde{\nabla}}\tilde{p} =\vec{0}\\
     &\tilde{\nabla}.\bm{\vec{\tilde{u}}} = 0\\
     &\bm{\vec{\tilde{u}}}\vec{\tilde{\nabla}}\tilde{T} = \dfrac{1}{Pe}\tilde{\nabla}^2\tilde{T} + \dfrac{Br}{Pe}\tilde{\eta}(\bm{\vec{\tilde{u}}}, \tilde{T})|\tilde{\gamma}(\bm{\vec{\tilde{u}}})|^2
\end{align*}
where $\bm{\vec{\tilde{u}}}=(\tilde{u}_x, \tilde{u}_y)^T$ is the velocity vector, $\tilde{p}$ is the pressure, $\tilde{T}$ is the temperature, $\tilde{\bar{\bar{\epsilon}}}$ is the strain-rate tensor, $\tilde{\eta}$ is the dynamic viscosity. Pe and Br are the dimensionless Péclet number and Brickman number, respectively.

We tackle an ill-posed configuration problem, i.e the boundary conditions of the problem are not completely defined, and we dispose of some measurements of the temperature. The goal is to infer the pressure, velocity, and temperature fields at all points in the domain. We note that here, no information on the pressure field is given, only its gradient in the PDE residual is. Thus the pressure is only identifiable up to a constant. As shown in \cite{nguyen2022physics}, only the information of the temperature is not sufficient to guarantee a unique solution for the velocity and the pressure fields, we take in addition the knowledge of velocity boundary conditions. The authors also showed that taking the collocation points on a finite-element (FE) mesh, which provides \textit{a priori} knowledge of high gradient location, improves significantly the accuracy of the prediction instead of taking the points randomly in the domain. However, to produce the FE mesh, expert physical knowledge is required. In this work, we show that FBOAL is able to infer automatically position for collocation points, and thus improve the performance of PINNs without the need for any FE mesh.

For the architecture of PINNs, we use a feedforward network with $5$ hidden layers and $100$ neurons per layer. To minimize the cost function, we adopt Adam optimizer with the learning rate decay strategy. The results are obtained with 50,000 epochs with the learning rate $lr=10^{-3},$ 100,000 epochs with the learning rate $lr=10^{-4}$ and 150,000 epochs with $lr=10^{-5}$. For the training of PINNs, we suppose to dispose of the boundary condition for the velocity and $N_T=500$ points of measurements for the temperature that are randomly distributed inside the domain. The number of collocation points is fixed as $N_f=10,000$ points. For the training of FBOAL, in this primary investigation, we do not divide the domain. After every $k=25,000$ iterations, we add and remove $m= 2.5\%\times N_{pde}=250$ collocation points based on the PDE residuals. Again, we note that in this case, since there are four PDEs residuals, we divide the set of collocation points into four separated subsets whose cardinal equal to $N_{pde}/4=2,500$, and then add and remove $m/4$ points independently for each equation. 

\begin{figure}[H]
    \centering
    \subfloat[Random mesh]{{\includegraphics[width=5.4cm]{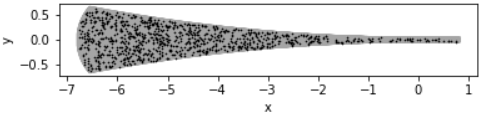} \label{calendre_colloc_rand}}}
    \subfloat[FE mesh]{{\includegraphics[width=5.4cm]{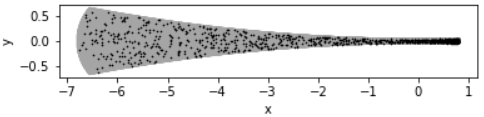} \label{calendre_colloc_fe}}}
    \subfloat[FBOAL]{{\includegraphics[width=5.4cm]{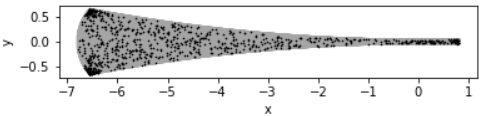} \label{calendre_colloc_boal}}}
    \caption{\textit{Calendering process:} Visualization of $1,000$ random collocation points in different scenarios}%
    \label{calendre_colloc_points}%
\end{figure}

\begin{figure}[H]
    \centering
    \subfloat[Random mesh]{{\includegraphics[width=5.3cm]{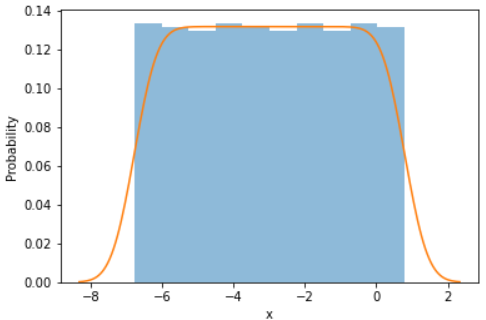} \label{burgers_calendre_rand}}}
    \subfloat[FE mesh]{{\includegraphics[width=5.2cm]{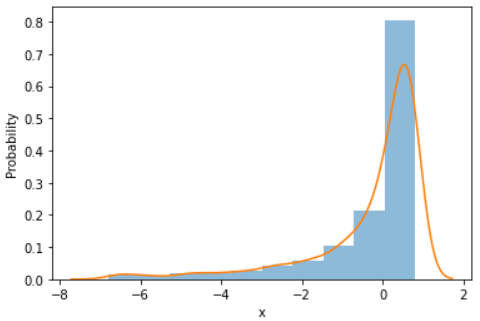} \label{burgers_calendre_fe}}}
    \subfloat[ FBOAL]{{\includegraphics[width=5.3cm]{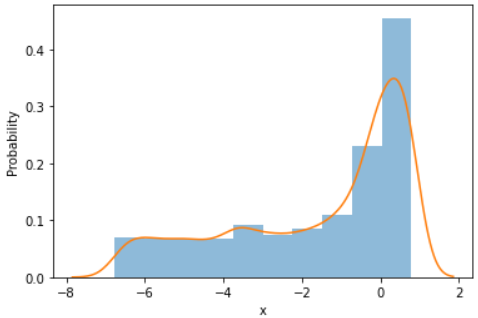} \label{burgers_calendre_boal}}}
    \caption{\textit{Calendering process: Density of points on the line y = 0}}%
    \label{burgers_calendre_dens}%
\end{figure}

\begin{figure}[H]
    \centering
    \subfloat[Random mesh]{{\includegraphics[width=5.3cm]{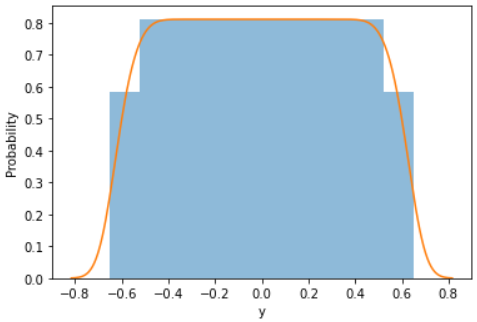} \label{burgers_calendre_rand_y}}}
    \subfloat[FE mesh]{{\includegraphics[width=5.2cm]{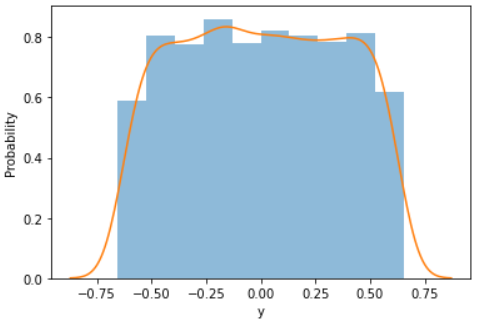} \label{burgers_calendre_fe_y}}}
    \subfloat[ FBOAL]{{\includegraphics[width=5.3cm]{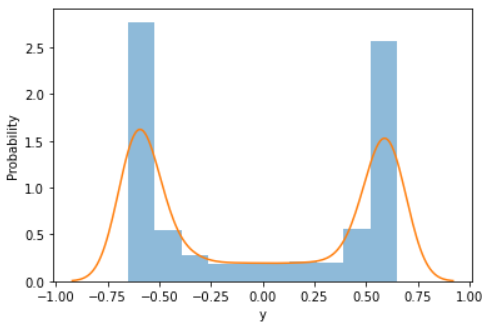} \label{burgers_calendre_boal_y}}}
    \caption{\textit{Calendering process: Density of points on the line x=-6.55}}%
    \label{burgers_calendre_dens_y}%
\end{figure}

Figure \ref{calendre_colloc_points} shows the visualization of collocation points on a random mesh, the FE mesh and after the training with FBOAL. Figure \ref{burgers_calendre_dens} and Figure \ref{burgers_calendre_dens_y} present the density of collocation points on the horizontal line $y=0$ and on the vertical line $x=-6.55$ (the line at the input of the calender where the contact between the rubber and the rolls begins). We see that, the main difference between the random mesh and the FE mesh is that the FE mesh is much finer at the output of the calender. However, with FBOAL, we see that the collocation points are not only added to the output of the calender, but also to the input where there is contact with the solid rolls.

\begin{figure}[H]
    \centering
    \subfloat[Reference solution]{{\includegraphics[width=4cm]{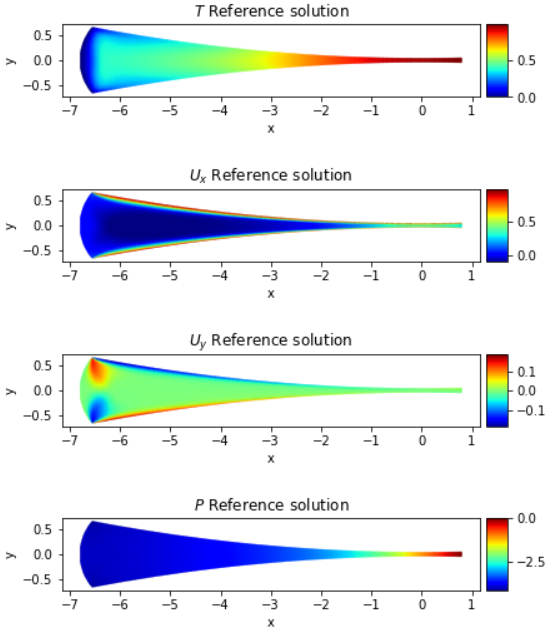} }}
    \subfloat[With random mesh]{{\includegraphics[width=4cm]{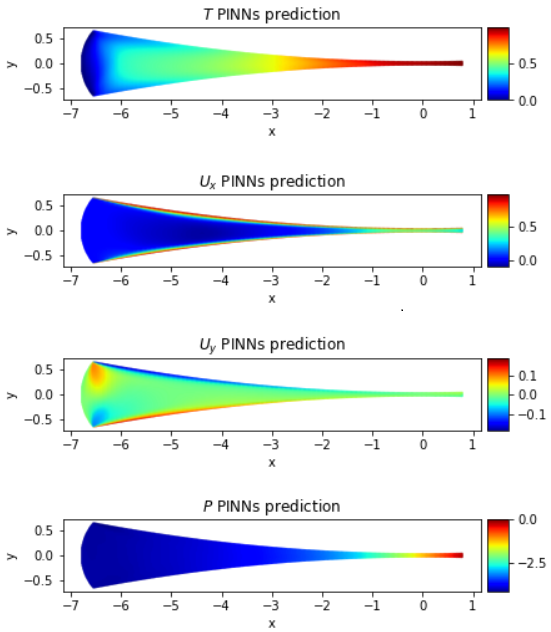} }}%
    \subfloat[With FE mesh]{{\includegraphics[width=4cm]{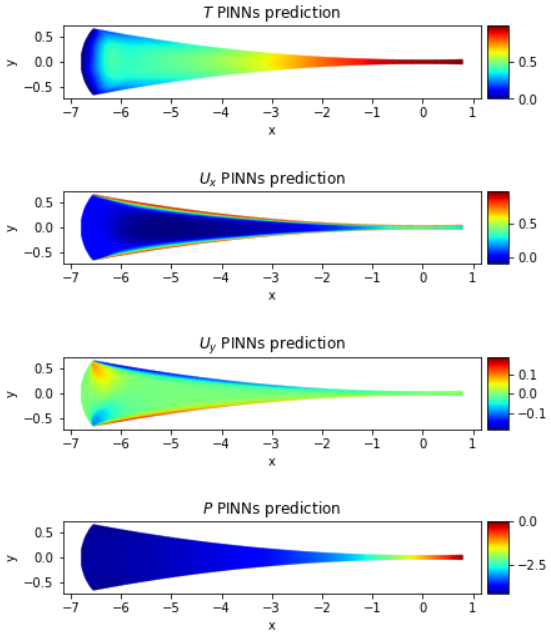} }}%
    \subfloat[With FBOAL]{{\includegraphics[width=4cm]{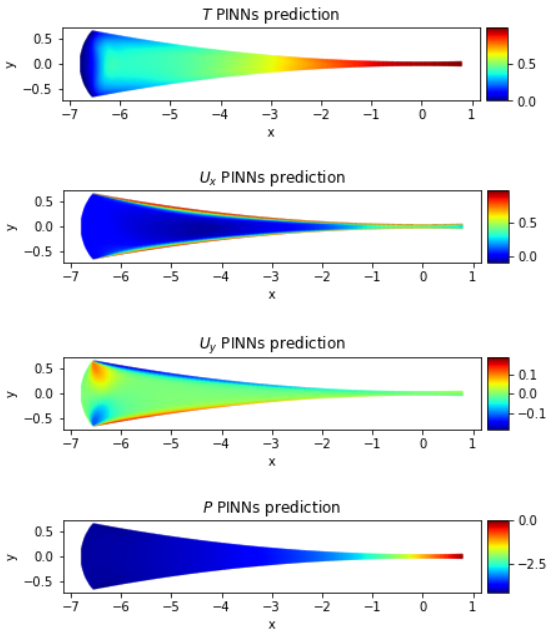} }}%
    \caption{\textit{Reference solution and PINNs prediction with different cases of collocation points.}}%
    \label{infer_sol_calendre}%
\end{figure}

\begin{table}[H]
\centering
\begin{tabular}{ |c|c|c|c|c| } 
\hline
 & $\epsilon_T$ & $\epsilon_{u_x}$ & $\epsilon_{u_y}$ & $\epsilon_P$\\
\hline
Random mesh &  13.6 $\pm$ 1.17 & 24.1 $\pm$ 3.77 & 15.6 $\pm$ 2.73 & 11.9 $\pm$ 1.33 \\
FE mesh & \textbf{8.54} $\pm$ \textbf{1.39} & \textbf{14.7} $\pm$ \textbf{2.49} & 18.0 $\pm$ 2.61 & \textbf{1.41} $\pm$ \textbf{0.17}\\
FBOAL & 10.5 $\pm$ 1.82 & 19.3 $\pm$ 1.85 & \textbf{9.63} $\pm$ \textbf{3.24} & 5.13 $\pm$ 0.72\\
\hline
\end{tabular}
\caption{\textit{Relative $\mathcal{L}^2$ errors between reference solution and PINNs prediction with different cases of collocation points.}}
\label{calendre_err}
\end{table}
To assess the performance of PINNs, we evaluate the errors of the prediction on a random mesh that contains $N=162,690$ points. To avoid any artificial high values of the error for fields very close to zero, we use a relative $\mathcal{L}^2$ error that divides the absolute error by the reference field amplitude, which is defined as follows:
\begin{align*}
    \epsilon_w = \dfrac{||w-\hat{w}||_2}{w_{max} - w_{min}}
\end{align*}
where $w$ denotes the reference simulated field of interest and $\hat{w}$ is the corresponding PINNs prediction. Table \ref{calendre_err} shows the performance of PINNs with different cases of collocation points in terms of relative $\mathcal{L}^2$ error. Figure \ref{infer_sol_calendre} gives more details about the visualization of the reference solution and the prediction of all models. We see that PINNs with FBOAL give better accuracy for all physical fields than classical PINNs with random collocation points. The classical PINNs with collocation points on the FE mesh still produce smaller errors for $T,u_x$ and $p$ than PINNs with FBOAL. However, for the vertical velocity component $u_y$, PINNs with FBOAL give the best prediction. This is because the solution for $u_y$ at the input of the calender has more complex behavior than at other zones. FBOAL is able to capture this complexity based on the PDEs residuals. The classical PINNs with the collocation points on FE mesh give the largest errors for $u_y$ since the FE mesh is very coarse at the input of the calender. With FBOAL, the algorithm is able to detect new zones which produce high errors for the PDEs residuals. 

\section{Conclusion}
In this paper, we introduced a Fixed-Budget Online Adaptive Learning (FBOAL) for the collocation points in PINNs that adds and removes points based on PDEs residuals on sub-domains while fixing the number of training points. Investigated the performance of FBOAL on Burgers equation in a non-parameterized problem (i.e. the viscosity is fixed) and a parameterized problem (i.e. the viscosity is varied) results showed that FBOAL provides better accuracy and fewer iterations than classical PINNs in both cases. Furthermore, FBOAL is able to relocate the collocation adaptively during the training so that the model is able to better capture the steep variation of the solution. The hyperparameters in FBOAL play important roles in the performance of FBOAL. The numerical results in various test cases show that the smaller values of $k$ give better accuracy for PINNs predictions, and the performance of PINNs is less sensitive to the values of $m$ and $d$ when $k$ is small. In all cases, we suggest starting with a small value of $m$ and $k$, and then increasing these values later to see whether the performance can be increased or not. Besides that, we also compared the performance of FBOAL with other existing methods of adaptive sampling for the collocation points such as Residual-based Adaptive Distribution (RAD) and Residual-based Adaptive Refinement with Distribution (RAR-D). It is shown that in most cases, FBOAL is able to provide a comparable or even better performance than other approaches in terms of accuracy and computation cost.  We then apply the strategy FBOAL to the rubber calendering process simulation. Motivated by the previous work of \cite{nguyen2022physics}, we aim to demonstrate that FBOAL can automatically relocate the collocation points at high-gradient locations. The results show that PINNs with FBOAL give better performance than classical PINNs with non-adaptive collocation points. Furthermore, FBOAL is able to add more points at the location of high-gradient (as the finite-element (FE) mesh) and at the zones where the solution for the vertical velocity component is hard to learn. For this component, PINNs with FBOAL give even better predictions than classical PINNs with the collocation points on FE mesh. This promising result demonstrates that FBOAL can help to provide \textit{a prior} knowledge of high-gradient locations and improve the conventional numerical solver in the construction of the mesh.

In this study, the collocation points have been relocated considering an error estimator based on the PDEs residuals. Taking ideas from the PDEs discretization community (\textit{e.g.} \cite{briffard2017contributions, belhamadia2019two}), it could be interesting to consider error estimators based on the solution approximation.

\bibliographystyle{plainnat} 
\bibliography{main} 
\appendix

\section{Burgers equation}\label{sec_annex_burgers}
\subsection{Non-parameterized problem}\label{annex_burgers_fix}

\begin{figure}[H]
    \centering
    \subfloat[When $d=50$, $k=1,000$]{{\includegraphics[width=5.5cm]{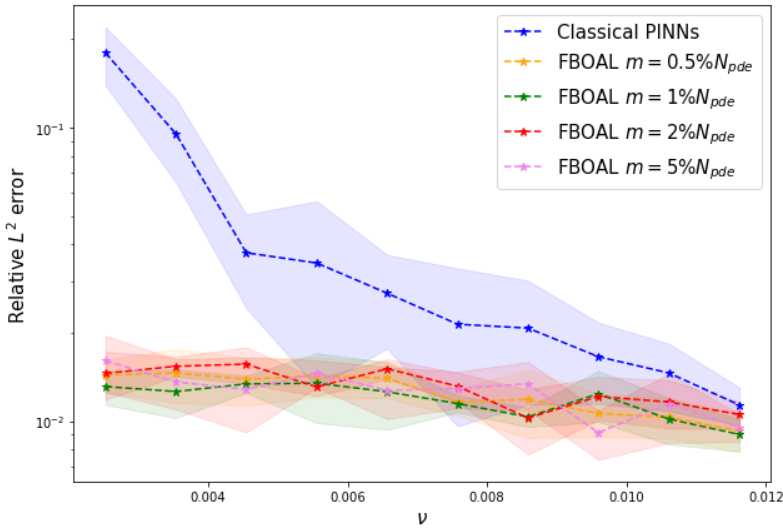} \label{burgers_fix_varym_1}}}
    \subfloat[When $d=50$, $k=2,000$]{{\includegraphics[width=5.4cm]{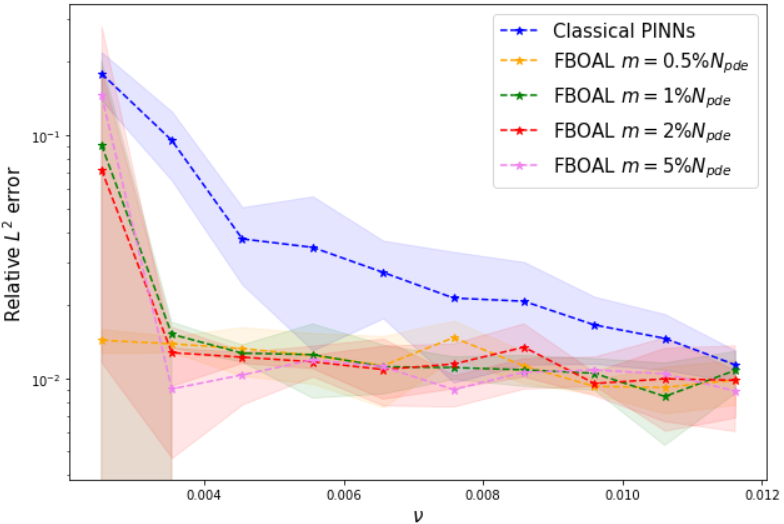} \label{burgers_fix_varym_2}}}
    \subfloat[When $d=50$, $k=5,000$]{{\includegraphics[width=5.4cm]{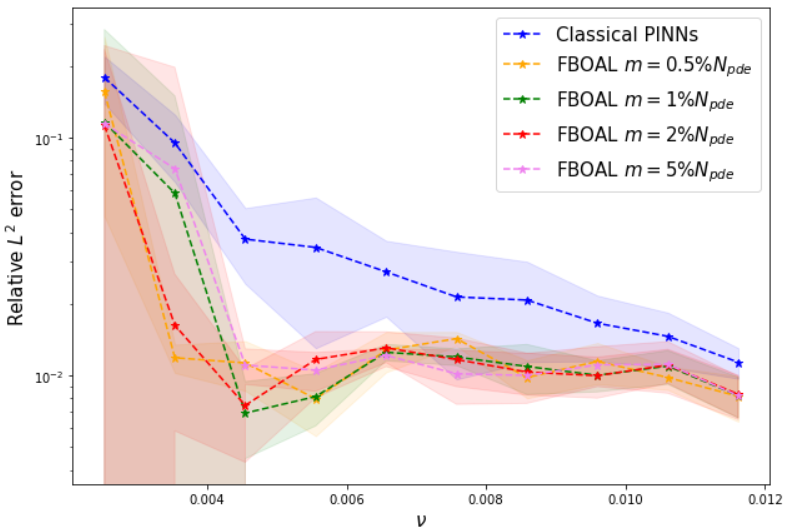} \label{burgers_fix_varym_3}}}
    \quad
    \subfloat[When $d=200$, $k=1,000$]{{\includegraphics[width=5.5cm]{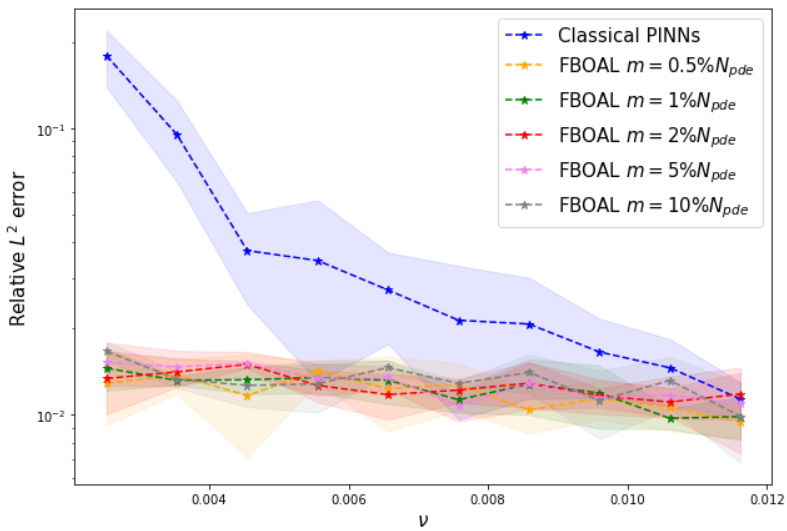}
    \label{burgers_fix_varym_4}}}
    \subfloat[When $d=200$, $k=2,000$]{{\includegraphics[width=5.5cm]{paper_boal_burgers_varym_err_ep2k_d200_decay.png}
    \label{burgers_fix_varym_5}}}
    \subfloat[When $d=200$, $k=5,000$]{{\includegraphics[width=5.5cm]{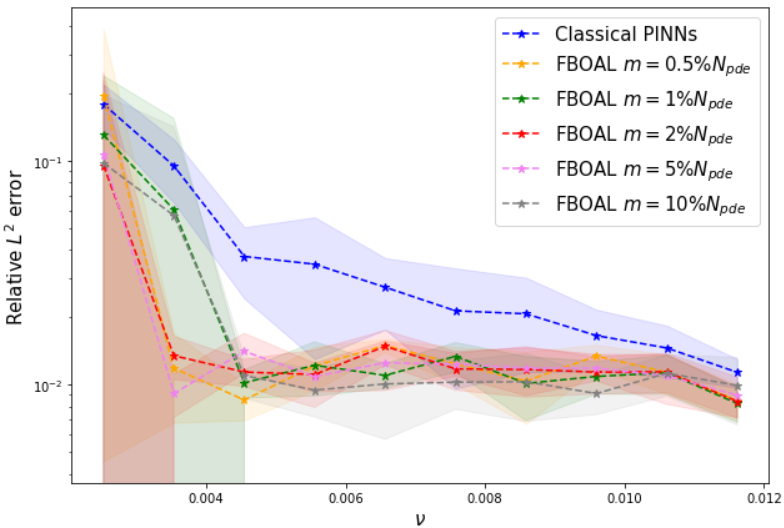}
    \label{burgers_fix_varym_6}}}
    \caption{\textit{Burgers equation: Performance in terms of accuracy of FBOAL when varying $m$.} The curves and shaded regions represent the geometric mean and one standard deviation of five runs.}%
    \label{burgers_boal_varym}%
\end{figure}


\begin{figure}[H]
    \centering
    \subfloat[When $d=50$, $m=0.5\%N_{pde}$]{{\includegraphics[width=5.5cm]{paper_boal_burgers_varyk_err_add5_d50_decay.png} \label{burgers_fix_varyk_1}}}
    \subfloat[When $d=50$, $m=1\%N_{pde}$]{{\includegraphics[width=5.4cm]{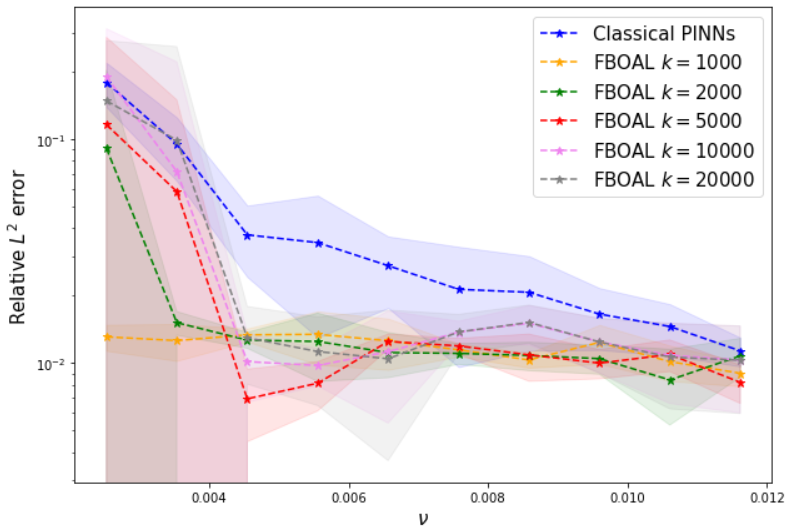} \label{burgers_fix_varyk_2}}}
    \subfloat[When $d=50$, $m=2\%N_{pde}$]{{\includegraphics[width=5.4cm]{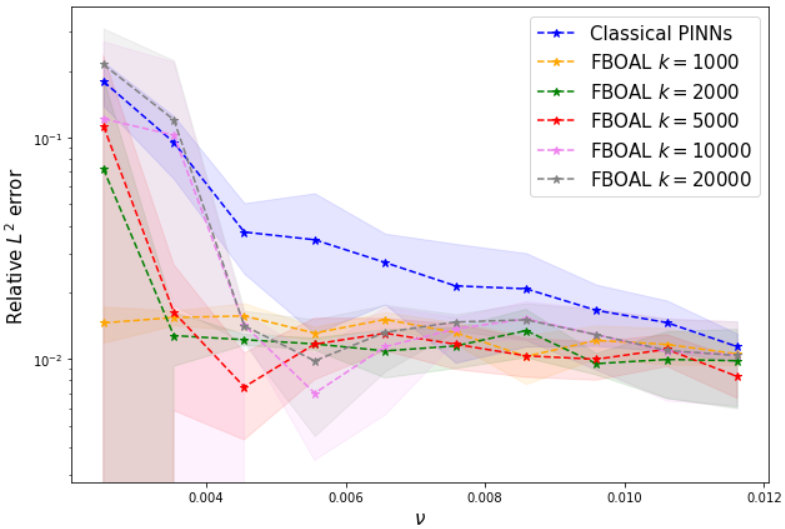} \label{burgers_fix_varyk_3}}}
    \quad
    \subfloat[When $d=200$, $m=0.5\%N_{pde}$]{{\includegraphics[width=5.5cm]{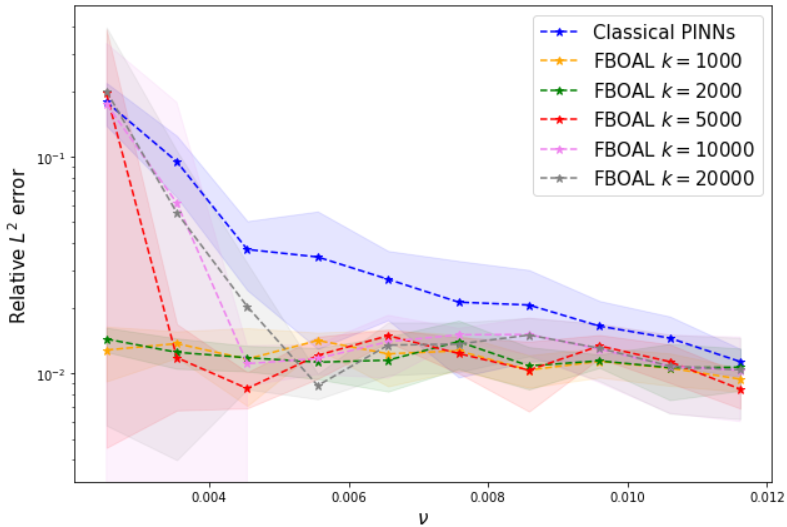}
    \label{burgers_fix_varyk_4}}}
    \subfloat[When $d=200$, $m=1\%N_{pde}$]{{\includegraphics[width=5.5cm]{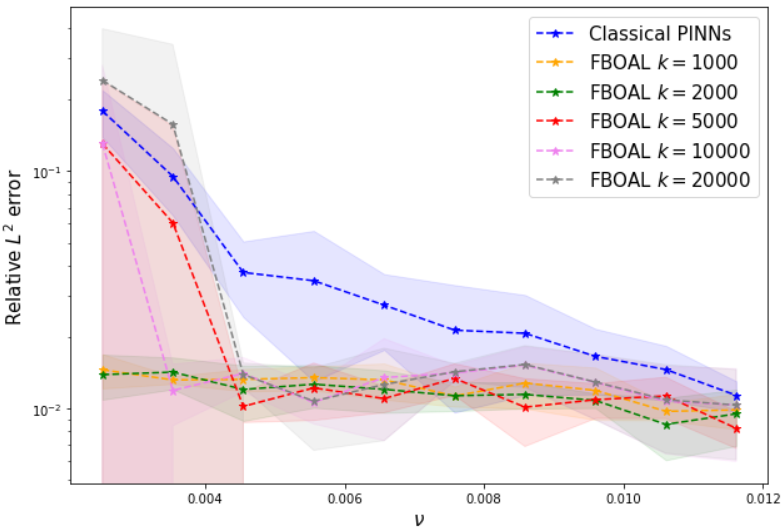}
    \label{burgers_fix_varyk_5}}}
    \subfloat[When $d=200$, $m=2\%N_{pde}$]{{\includegraphics[width=5.5cm]{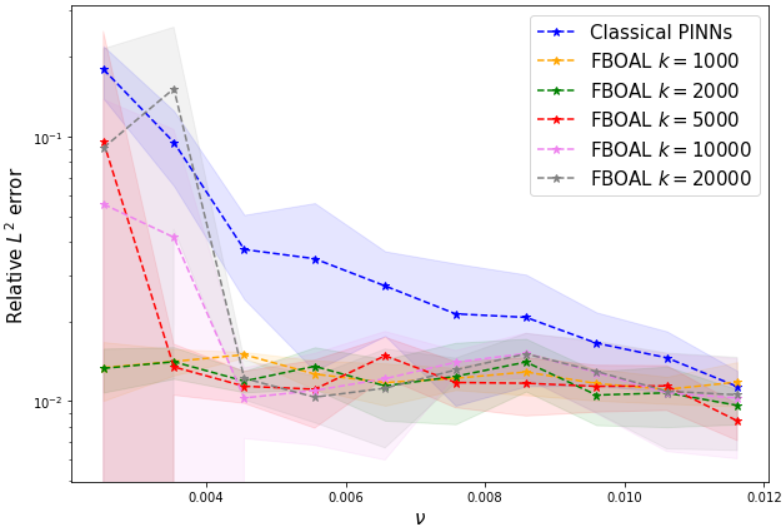}
    \label{burgers_fix_varyk_6}}}
    \quad
    \caption{\textit{Burgers equation: Performance in terms of accuracy of FBOAL when varying $k$.} The curves and shaded regions represent the geometric mean and one standard deviation of five runs.}%
    \label{burgers_boal_varyk}%
\end{figure}


\begin{figure}[H]
    \centering
    \subfloat[When $m=0.5\%N_{pde}$, $k=1,000$]{{\includegraphics[width=5.5cm]{paper_boal_burgers_varyd_err_add5_ep1k_decay.png} \label{burgers_fix_varyd_1}}}
    \subfloat[When $m=1\%N_{pde}$, $k=1,000$]{{\includegraphics[width=5.4cm]{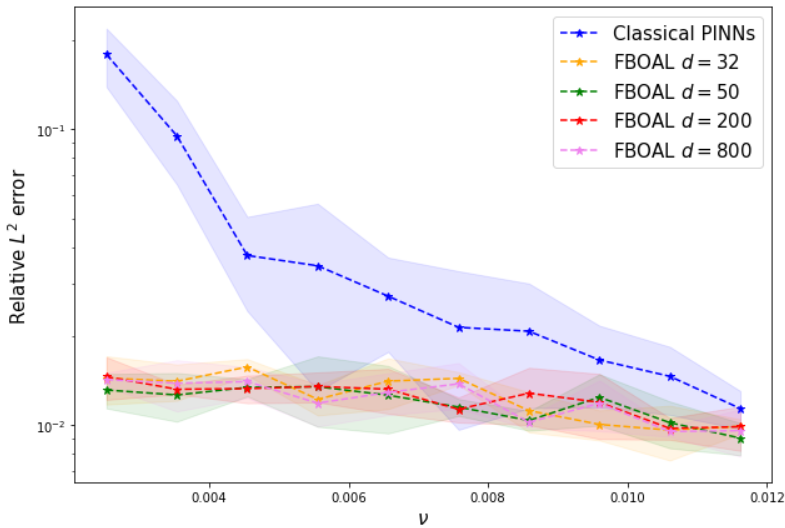} \label{burgers_fix_varyd_2}}}
    \subfloat[When $m=2\%N_{pde}$, $k=1,000$]{{\includegraphics[width=5.4cm]{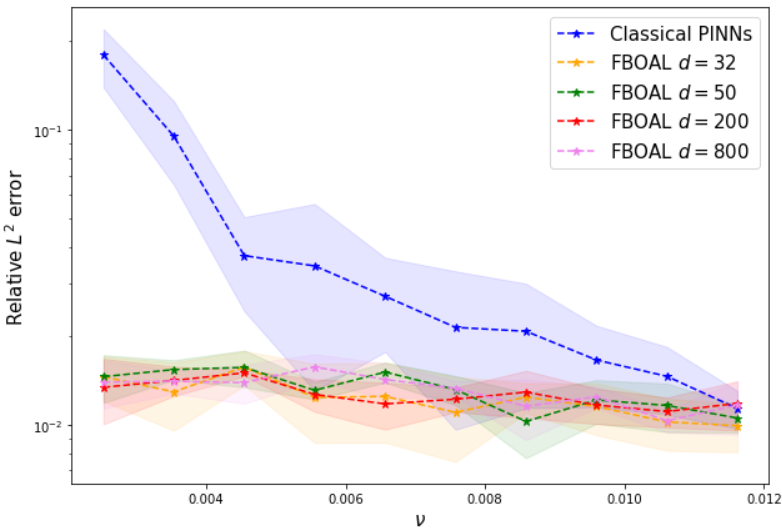} \label{burgers_fix_varyd_3}}}
    \quad
    \subfloat[When $m=0.5\%N_{pde}$, $k=2,000$]{{\includegraphics[width=5.5cm]{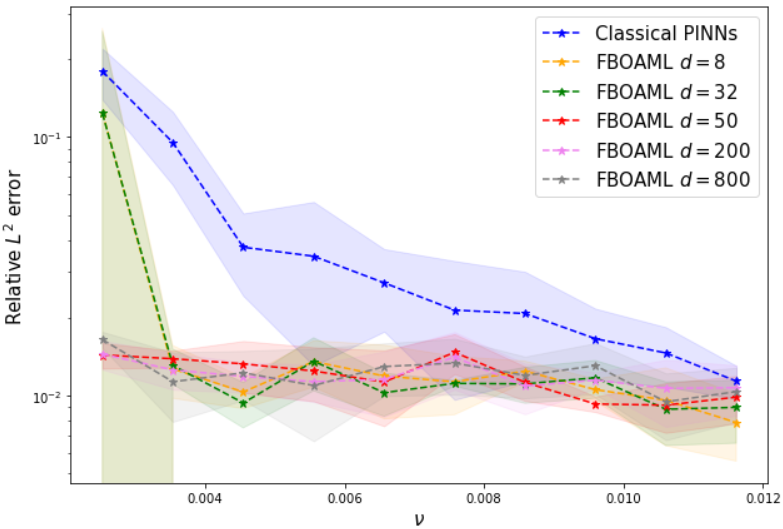}
    \label{burgers_fix_varyd_4}}}
    \subfloat[When $m=1\%N_{pde}$, $k=2,000$]{{\includegraphics[width=5.5cm]{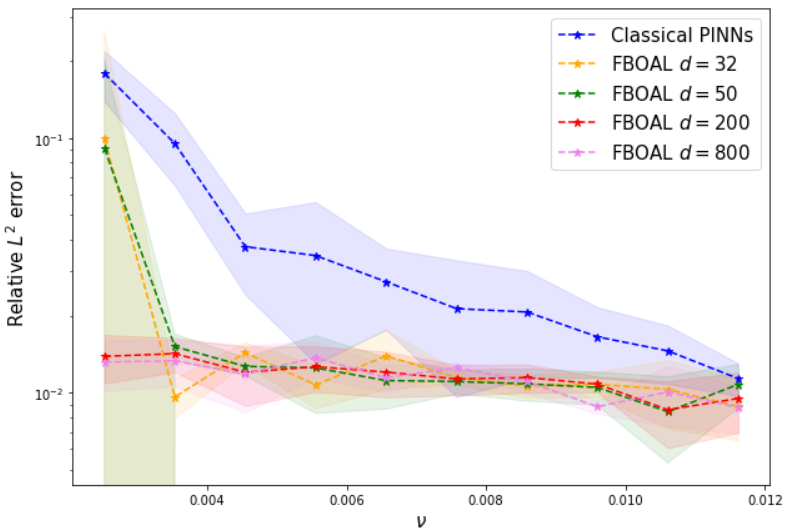}
    \label{burgers_fix_varyd_5}}}
    \subfloat[When $m=2\%N_{pde}$, $k=2,000$]{{\includegraphics[width=5.5cm]{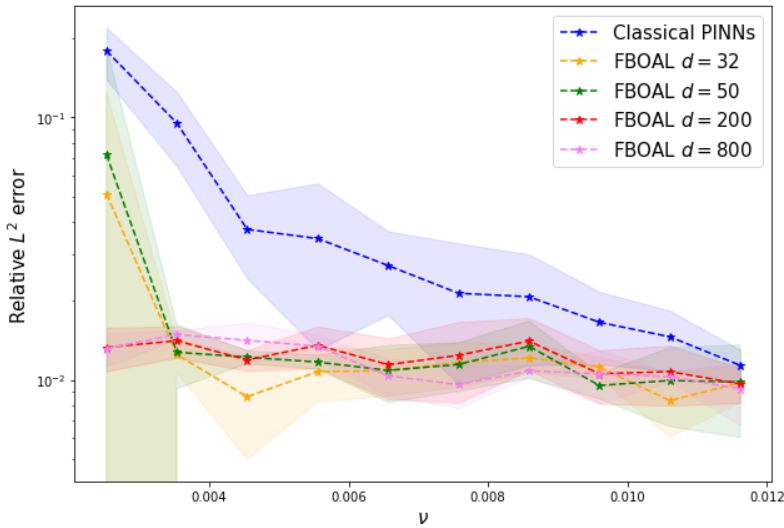}
    \label{burgers_fix_varyd_6}}}
    \quad
    \caption{\textit{Burgers equation: Performance in terms of accuracy of FBOAL when varying $d$.} The curves and shaded regions represent the geometric mean and one standard deviation of five runs.}%
    \label{burgers_boal_varyd}%
\end{figure}


\subsection{Parameterized problem}\label{annex_burgers_param}

\begin{figure}[H]
    \centering
    \subfloat[When $d=50$, $k=1,000$]{{\includegraphics[width=5.5cm]{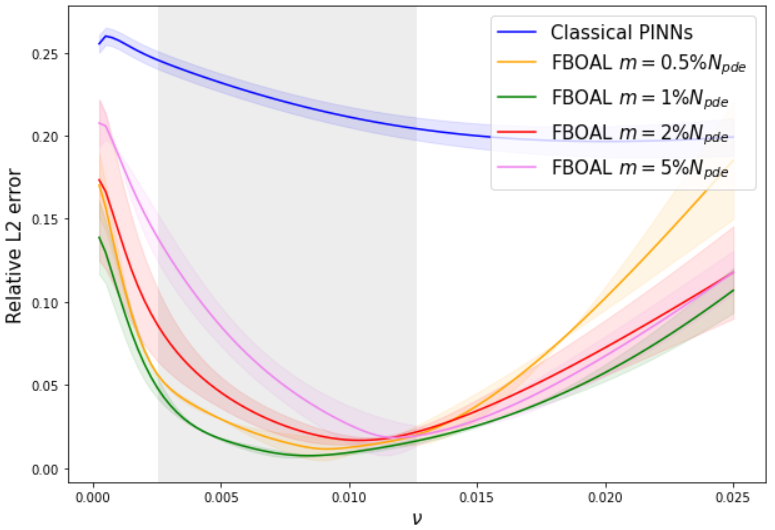} \label{burgers_param_varym_1}}}
    \subfloat[When $d=50$, $k=2,000$]{{\includegraphics[width=5.4cm]{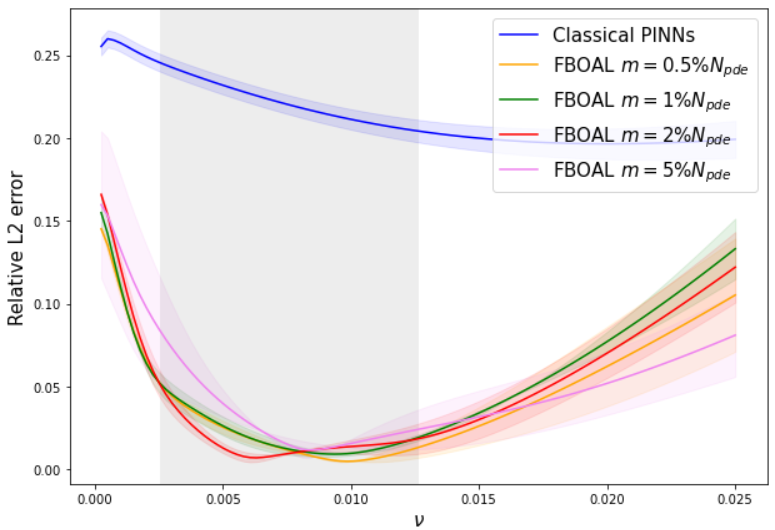} \label{burgers_param_varym_2}}}
    \subfloat[When $d=50$, $k=5,000$]{{\includegraphics[width=5.4cm]{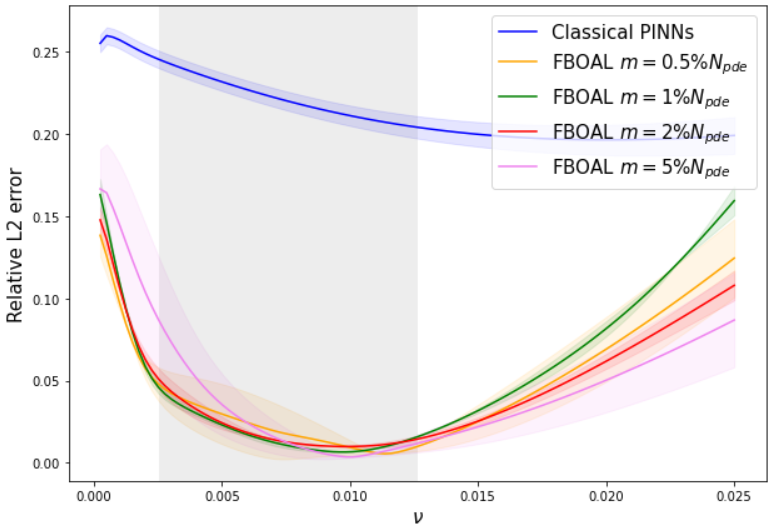} \label{burgers_param_varym_3}}}
    \quad
    \subfloat[When $d=200$, $k=1,000$]{{\includegraphics[width=5.5cm]{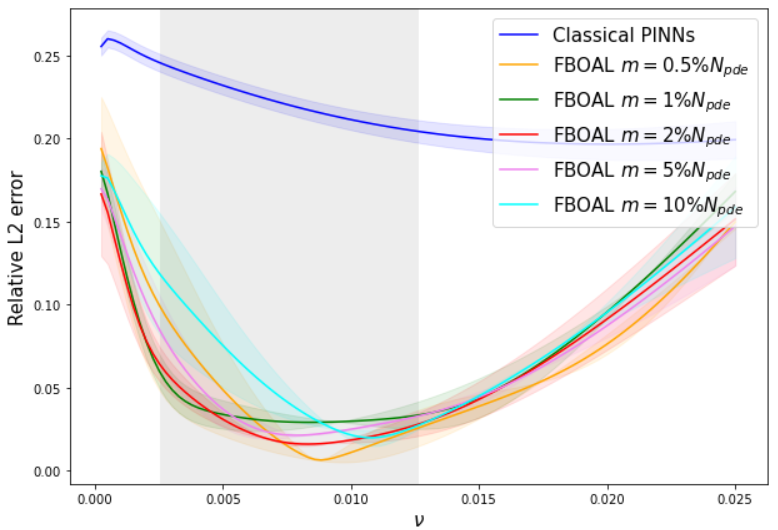}
    \label{burgers_param_varym_4}}}
    \subfloat[When $d=200$, $k=2,000$]{{\includegraphics[width=5.5cm]{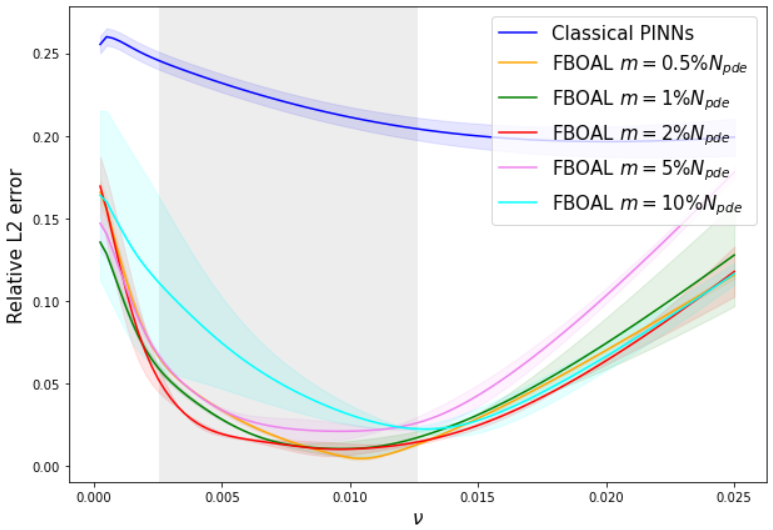}
    \label{burgers_param_varym_5}}}
    \subfloat[When $d=200$, $k=5,000$]{{\includegraphics[width=5.5cm]{paper_pinns_boal_burgers_params_varym_ep5k_s01.png}
    \label{burgers_param_varym_6}}}
    \caption{\textit{Burgers equation: Performance in terms of accuracy of FBOAL when varying $m$.} The curves and shaded regions represent the geometric mean and one standard deviation of five runs.}%
    \label{burgers_boal_param_varym}%
\end{figure}


\begin{figure}[H]
    \centering
    \subfloat[When $d=50$, $m=0.5\%N_{pde}$]{{\includegraphics[width=5.5cm]{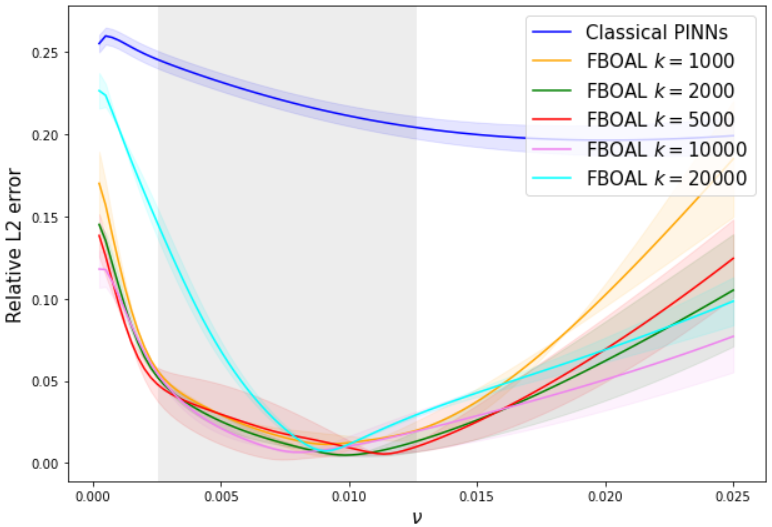} \label{burgers_param_varyk_1}}}
    \subfloat[When $d=50$, $m=1\%N_{pde}$]{{\includegraphics[width=5.4cm]{paper_pinns_boal_burgers_params_varyk_add400_s02.png} \label{burgers_param_varyk_2}}}
    \subfloat[When $d=50$, $m=2\%N_{pde}$]{{\includegraphics[width=5.4cm]{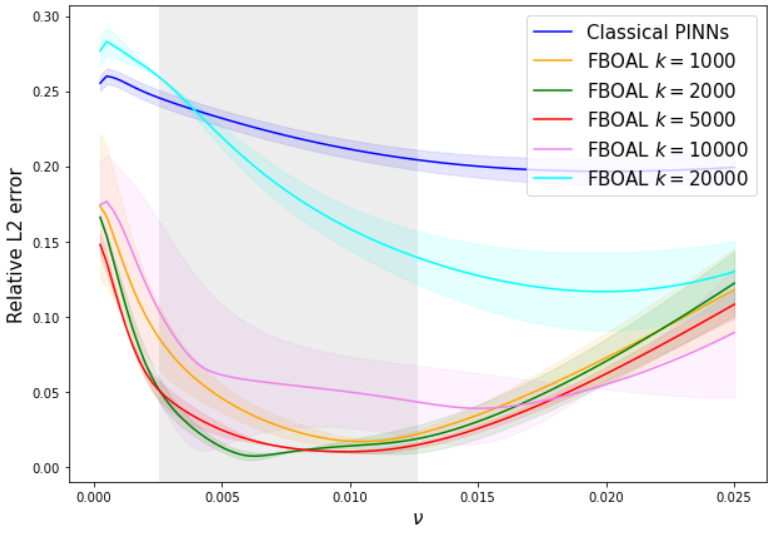} \label{burgers_param_varyk_3}}}
    \quad
    \subfloat[When $d=200$, $m=0.5\%N_{pde}$]{{\includegraphics[width=5.5cm]{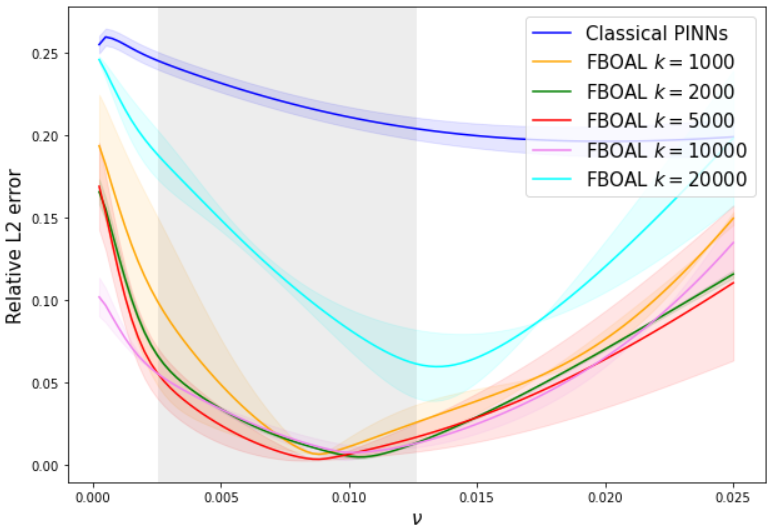}
    \label{burgers_param_varyk_4}}}
    \subfloat[When $d=200$, $m=1\%N_{pde}$]{{\includegraphics[width=5.5cm]{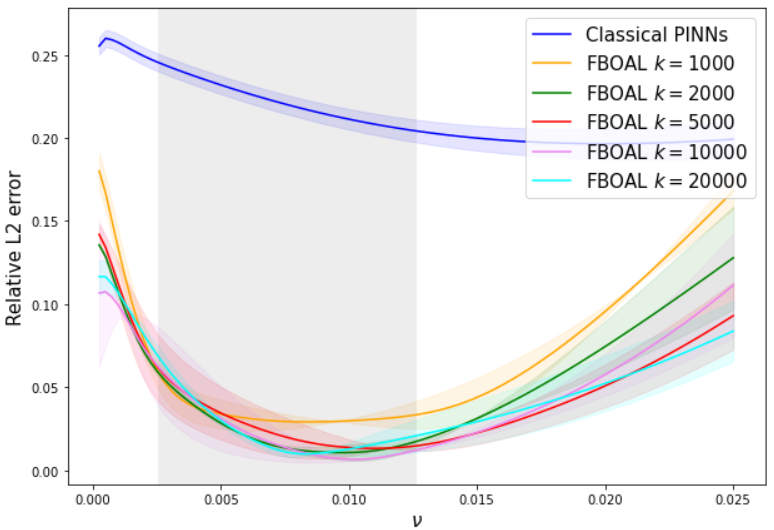}
    \label{burgers_param_varyk_5}}}
    \subfloat[When $d=200$, $m=2\%N_{pde}$]{{\includegraphics[width=5.5cm]{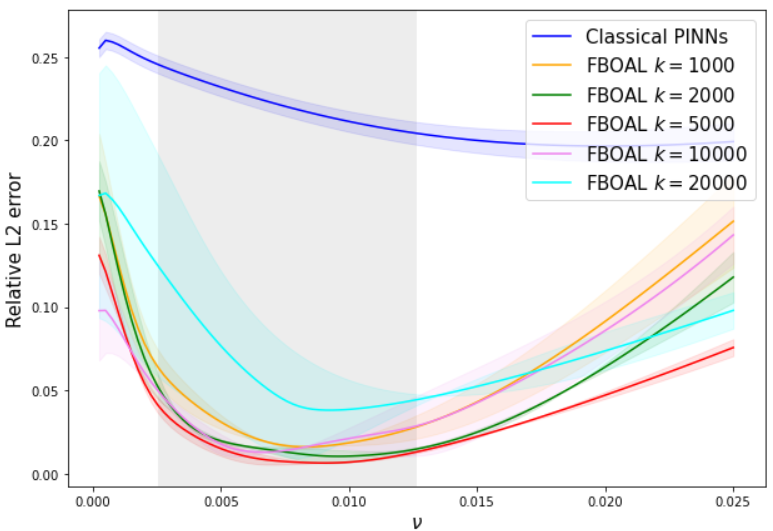}
    \label{burgers_param_varyk_6}}}
    \quad
    \caption{\textit{Burgers equation: Performance in terms of accuracy of FBOAL when varying $k$.} The curves and shaded regions represent the geometric mean and one standard deviation of five runs.}%
    \label{burgers_boal_param_varyk}%
\end{figure}

\begin{figure}[H]
    \centering
    \subfloat[When $m=0.5\%N_{pde}$, $k=2,000$]{{\includegraphics[width=5.5cm]{paper_pinns_boal_burgers_params_varyd_add200_ep2k.png} \label{burgers_param_varyd_1}}}
    \subfloat[When $m=1\%N_{pde}$, $k=2,000$]{{\includegraphics[width=5.4cm]{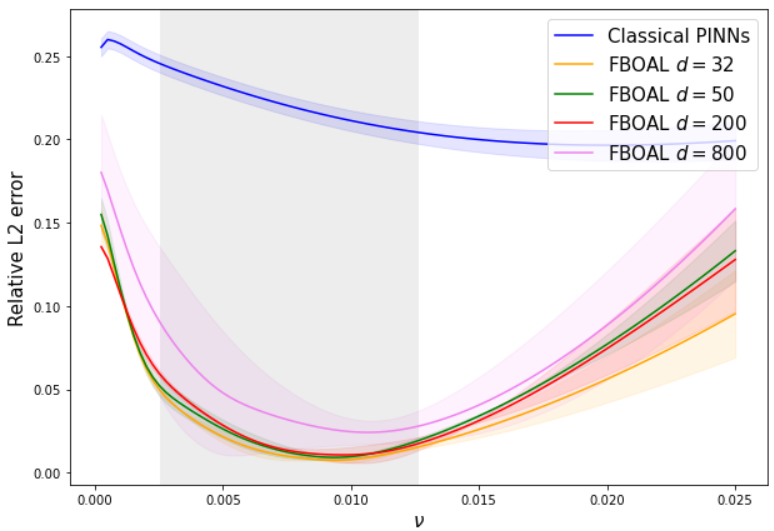} \label{burgers_param_varyd_2}}}
    \subfloat[When $m=2\%N_{pde}$, $k=2,000$]{{\includegraphics[width=5.4cm]{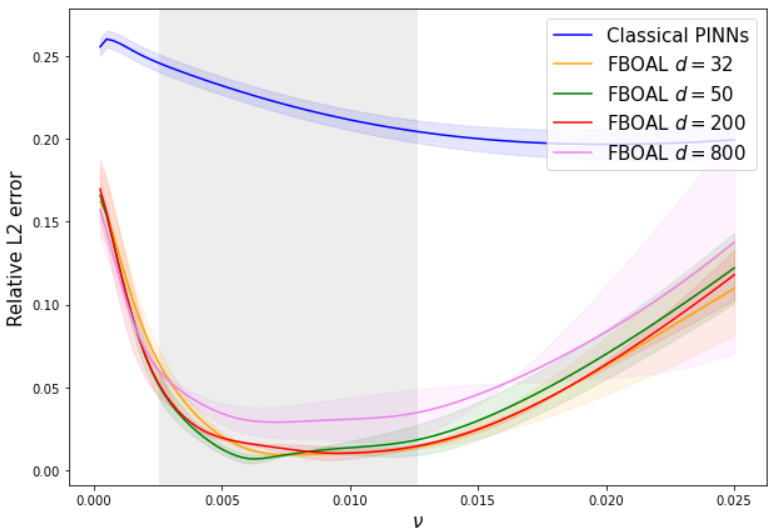} \label{burgers_param_varyd_3}}}
    \quad
    \subfloat[When $m=0.5\%N_{pde}$, $k=5,000$]{{\includegraphics[width=5.5cm]{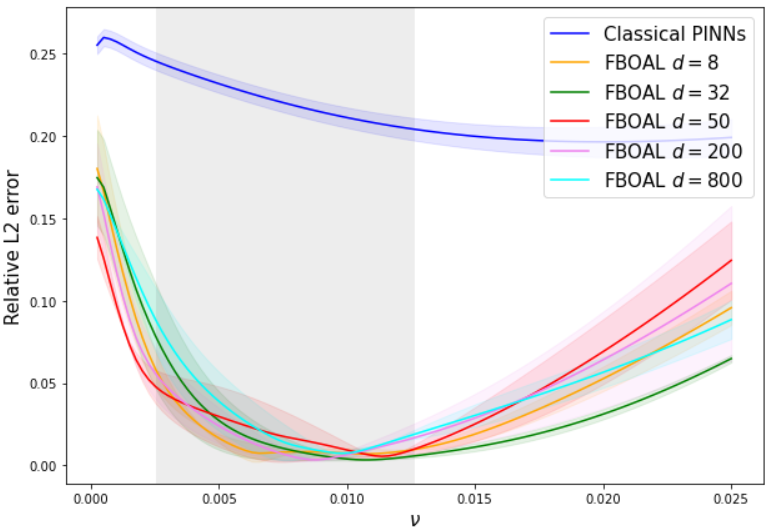}
    \label{burgers_param_varyd_4}}}
    \subfloat[When $m=1\%N_{pde}$, $k=5,000$]{{\includegraphics[width=5.5cm]{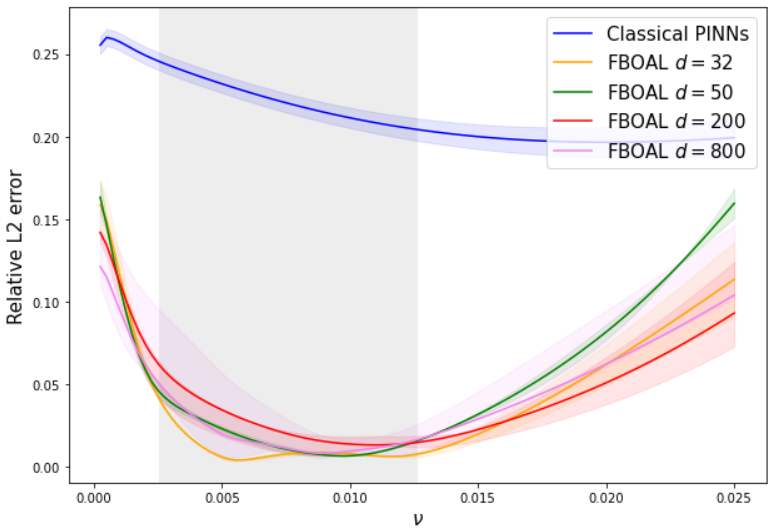}
    \label{burgers_param_varyd_5}}}
    \subfloat[When $m=2\%N_{pde}$, $k=5,000$]{{\includegraphics[width=5.5cm]{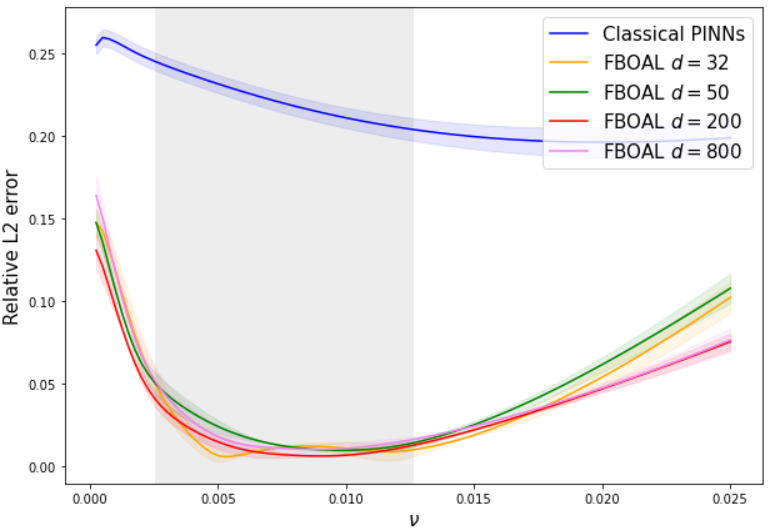}
    \label{burgers_param_varyd_6}}}
    \quad
    \caption{\textit{Burgers equation: Performance in terms of accuracy of FBOAL when varying $d$.} The curves and shaded regions represent the geometric mean and one standard deviation of five runs.}%
    \label{burgers_boal_param_varyd}%
\end{figure}

\section{Wave equation}\label{sec_wave}

\begin{figure}[H]
    \centering
    \subfloat[For $c^2=1$]{{\includegraphics[width=5.5cm]{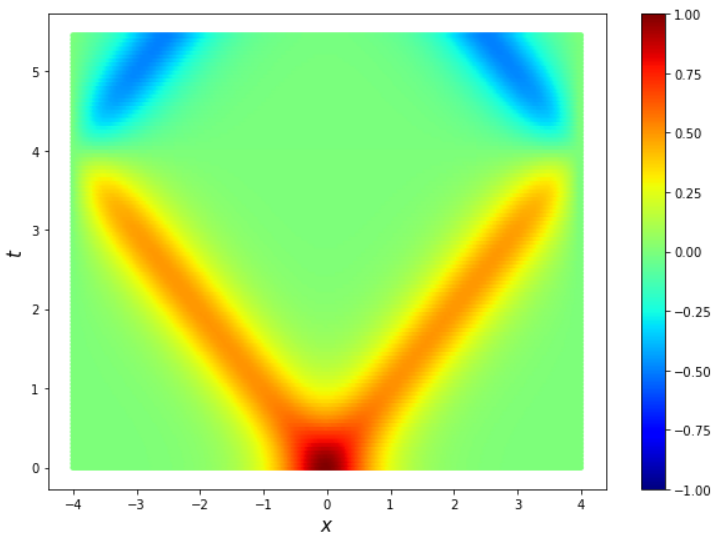} \label{wave_sol_1}}}
    \subfloat[For $c^2=2$]{{\includegraphics[width=5.5cm]{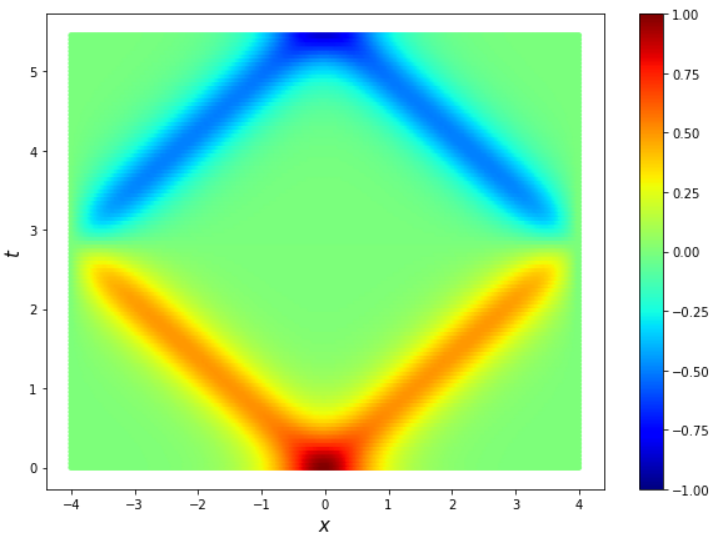} \label{wave_sol_2}}}
    \subfloat[For $c^2=3$]{{\includegraphics[width=5.5cm]{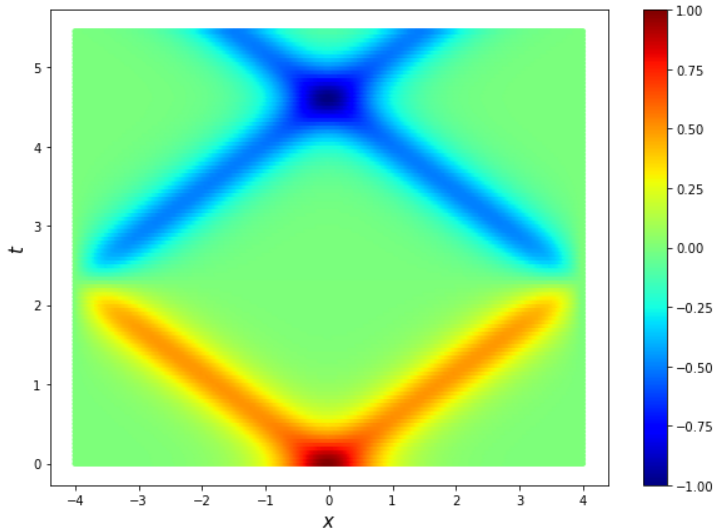} \label{wave_sol_3}}}
    \caption{\textit{Wave equation: Reference solution for different values of $c^2$.}}%
    \label{wave_sol}%
\end{figure}

We consider the following 1D wave equation, which is similar to the wave equation considered in \cite{peng2022rang}:
\begin{align*}
\begin{cases}
  u_{tt} + c^2u_{xx} = 0 \quad \text{for } x\in [-l,l], t\in[0, T] \\
  u(x,0) = \dfrac{1}{\cosh(2x)} - \dfrac{1}{2\cosh(2(x-2l))}+\dfrac{1}{2\cosh(2(x+2l))}\\
  u(-l,t) = u(l,t) = 0\\
  u_t(x,0) = 0
\end{cases}
\end{align*}
with $l=4$ and $T=5.5$. The exact solution is given as:
\begin{align*}
    u(x,t) = \dfrac{1}{2\cosh(2(x+ct))} - \dfrac{1}{2\cosh(2(x-2l+ct))} + \dfrac{1}{2\cosh(2(x-ct))} - \dfrac{1}{2\cosh(2(x+2l-ct))}
\end{align*}
Figure \ref{wave_sol} gives the visualization of the exact solution for different values of the parameter $c^2$.

\subsection{Non-parameterized problem}
We illustrate the performance of adaptive sampling approaches in a context where $c^2$ is fixed. We take $11$ values of $c^2\in[1, 3]$. For each $c^2$, we compare the performance of classical PINNs, PINNs with RAD, RAR-D, and PINNs with FBOAL. For the training of PINNs, we take $N_{pde}=1024$ collocation points that are initialized equidistantly inside the domain. We take a testing set of reference solutions on a $10\times 10$ equidistant spatio-temporal mesh and stop the training when either the number of iterations surpasses $K=300,000$ or the relative $\mathcal{L}^2$ error between PINNs prediction and the testing reference solution is smaller than the threshold $s=0.005$.

\begin{figure}[H]
    \centering
    \subfloat[Relative $\mathcal{L}^2$ error]{{\includegraphics[width=5.5cm]{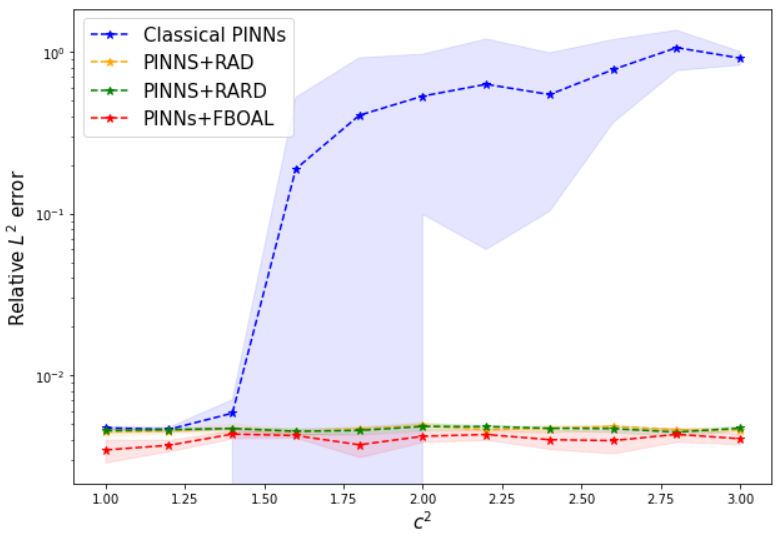} \label{wave_fix_err}}}
    \subfloat[Number of training iterations]{{\includegraphics[width=5.5cm]{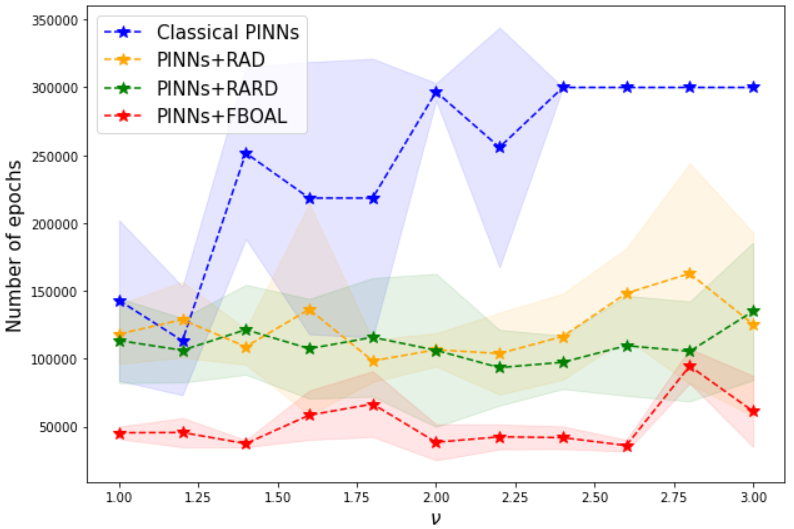} \label{wave_fix_nb}}}
    \subfloat[Cost function for $c^2=3$]{{\includegraphics[width=5.5cm]{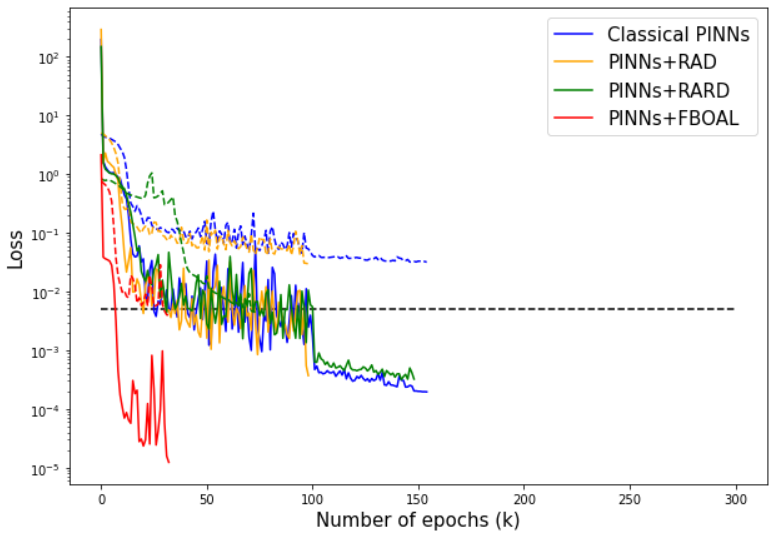} \label{wave_fix_cost}}}
    \caption{\textit{Wave equation: Comparison of classical PINNs and PINNs with adaptive sampling approaches.} The curves and shaded regions represent the geometric mean and one standard deviation of five runs.}%
    \label{wave_fix_compare}%
\end{figure}

\begin{figure}[H]
    \centering
    \subfloat[For $c^2=1$]{{\includegraphics[width=5.5cm]{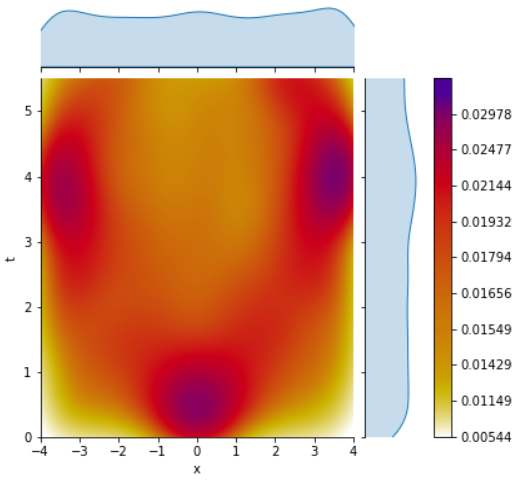} \label{wave_fix_dnu0}}}
    \subfloat[For $c^2=2$]{{\includegraphics[width=5.5cm]{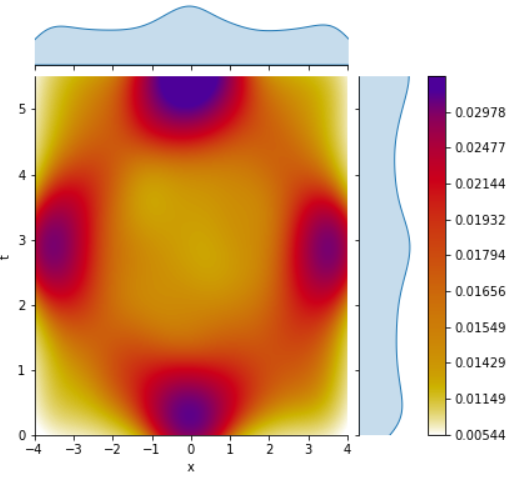} \label{wave_fix_dnu1}}}
    \subfloat[For $c^2=3$]{{\includegraphics[width=5.5cm]{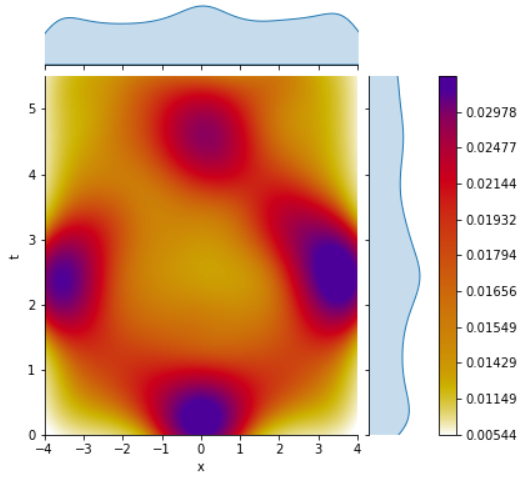} \label{wave_fix_dnu2}}}
    \caption{\textit{Wave equation: Density of collocation points after the training with FBOAL.}}%
    \label{wave_fix_dens}%
\end{figure}

\begin{figure}[H]
    \centering
    \subfloat[When $k=1,000$, $d=196$]{{\includegraphics[width=5.5cm]{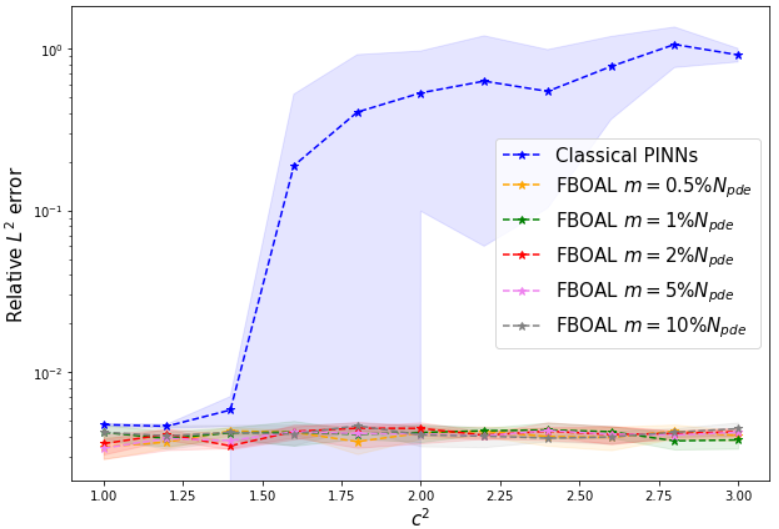} \label{wave_fix_varym}}}
    \subfloat[When $m=1\%N_{pde}$, $d=196$]{{\includegraphics[width=5.5cm]{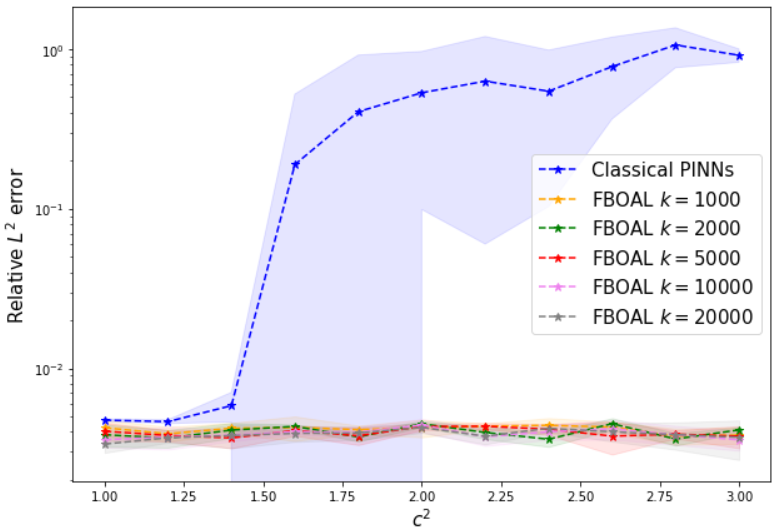} \label{wave_fix_varyk}}}
    \subfloat[When $k=2,000$, $m=1\%N_{pde}$]{{\includegraphics[width=5.5cm]{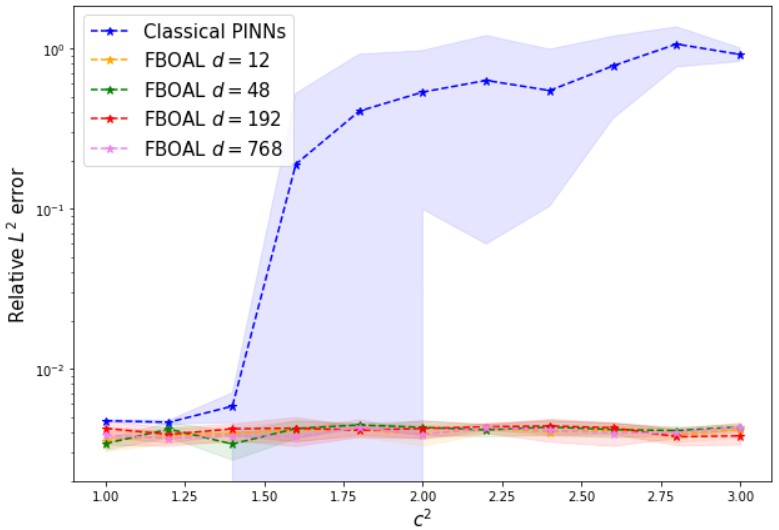} \label{wave_fix_varyd}}}
    \caption{\textit{Wave equation: Performance of FBOAL when varying $m$, $k$, and $d$.}}%
    \label{wave_fix_vary}%
\end{figure}

\subsection{Parameterized problem}
We illustrate the performance of PINNs in a parameterized problem where $c^2$ can be varied. In this case, $c^2$ is also considered as an input of PINNs. For the training, we take $41$ values of $\nu\in[1,3]$. For each $c^2$, we initialize 1024 collocation points, which leads to $N_{pde}=1024 \times 41= 41,984$ collocation points in total. We take a testing set of reference solutions on a $10\times 10$ equidistant spatio-temporal mesh and stop the training when either the number of iterations surpasses $K=2 \times 10^6$ or the sum of relative $\mathcal{L}^2$ error between PINNs prediction and the testing reference solution of all learning values of $\nu$ is smaller than the threshold $s=0.005\times 41=0.205$.

\begin{figure}[H]
    \centering
    \subfloat[Relative $\mathcal{L}^2$ error]{{\includegraphics[width=5.5cm]{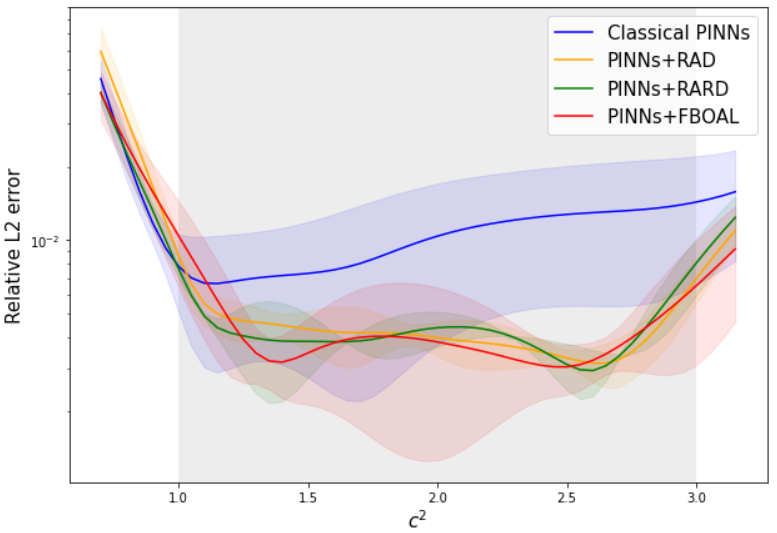} \label{wave_params_err}}}
    \subfloat[Number of collocation points]{{\includegraphics[width=5.5cm]{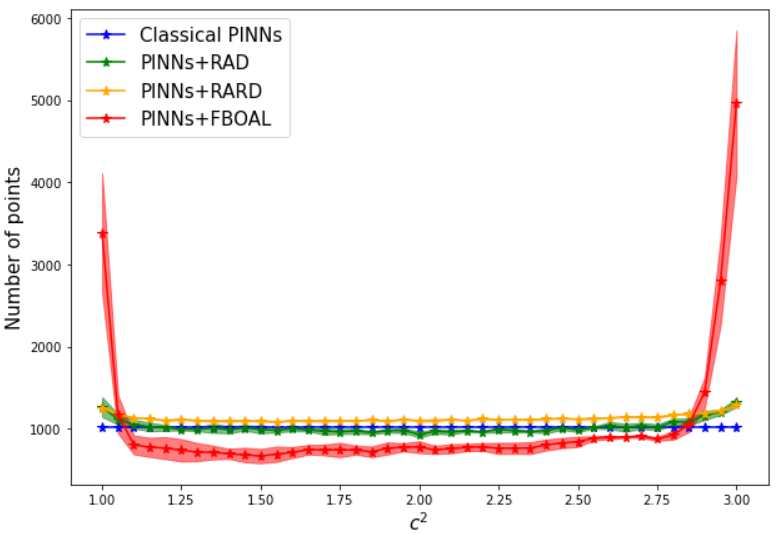} \label{wave_params_nb}}}
    \subfloat[Cost function during the training]{{\includegraphics[width=5.5cm]{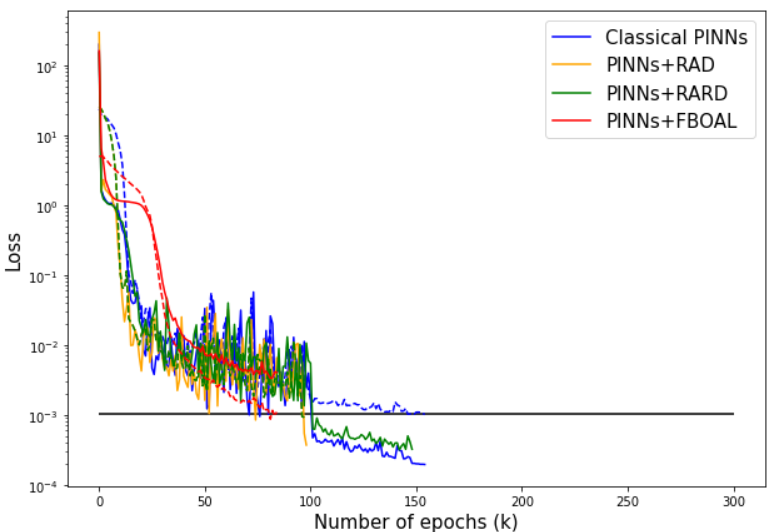} \label{wave_params_cost}}}
    \caption{\textit{Wave equation: comparison of classical PINNs and PINNs with adaptive sampling approaches.} The zone in gray is the learning interval for $c^2$ (interpolation zone). The curves and shaded regions represent the geometric mean and one standard deviation of five runs.}%
    \label{wave_params_compare}%
\end{figure}

\begin{figure}[H]
    \centering
    \subfloat[For $c^2=1$]{{\includegraphics[width=5.5cm]{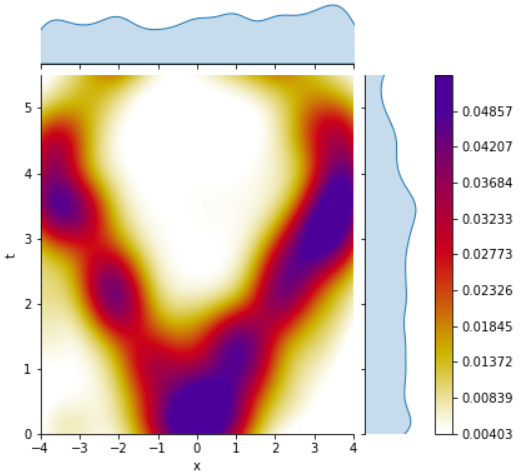} \label{wave_params_dnu0}}}
    \subfloat[For $c^2=2$]{{\includegraphics[width=5.5cm]{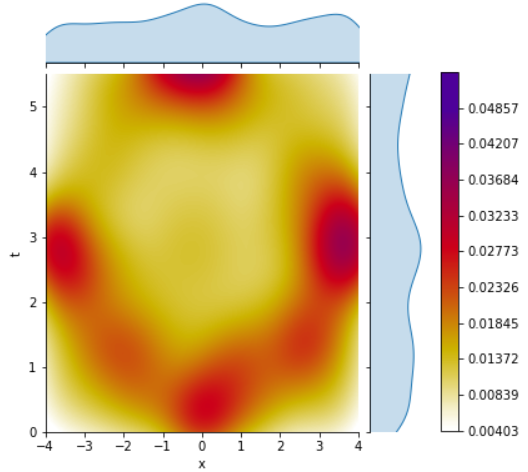} \label{wave_params_dnu1}}}
    \subfloat[For $c^2=3$]{{\includegraphics[width=5.5cm]{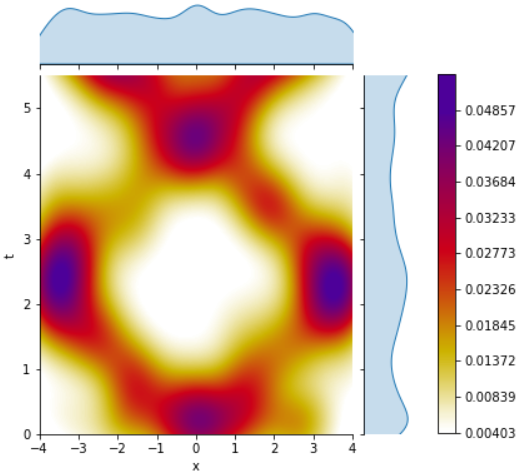} \label{wave_params_dnu2}}}
    \caption{\textit{Wave equation: Density of collocation points after the training by FBOAL.}}%
    \label{wave_params_dens}%
\end{figure}

\begin{figure}[H]
    \centering
    \subfloat[When $k=5,000$, $d=196$]{{\includegraphics[width=5.5cm]{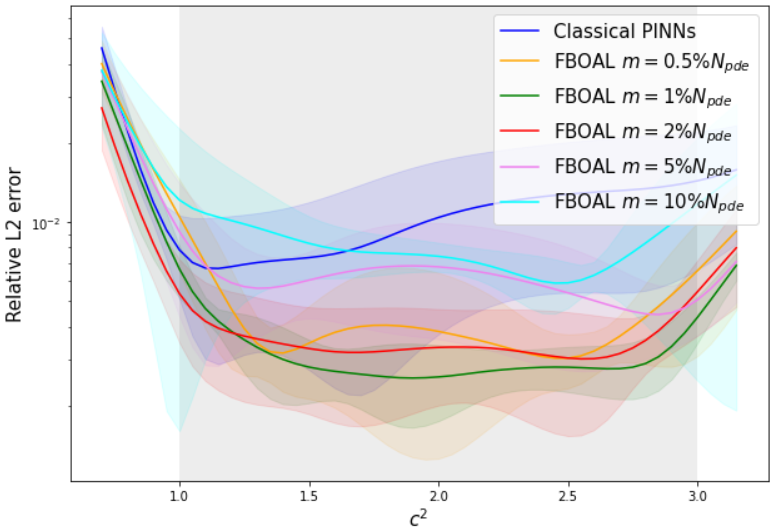} \label{wave_params_varym}}}
    \subfloat[When $m=1\%N_{pde}$, $d=196$]{{\includegraphics[width=5.5cm]{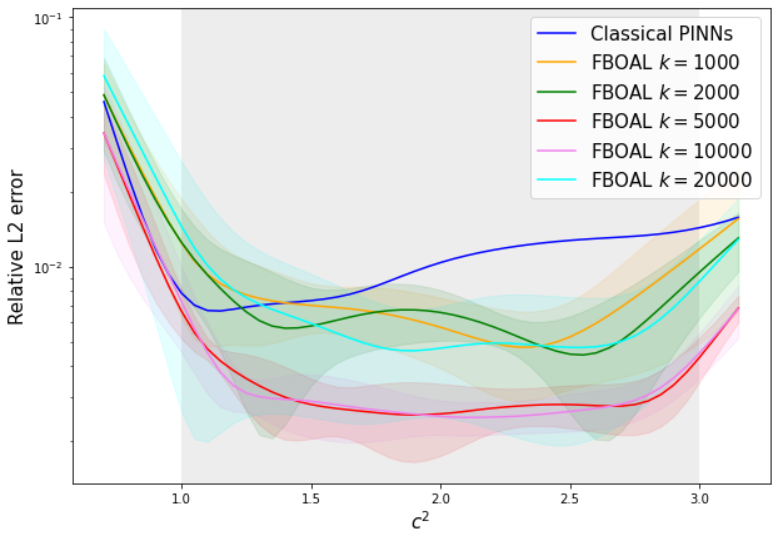} \label{wave_params_varyk}}}
    \subfloat[When $k=5,000$, $m=1\%N_{pde}$]{{\includegraphics[width=5.5cm]{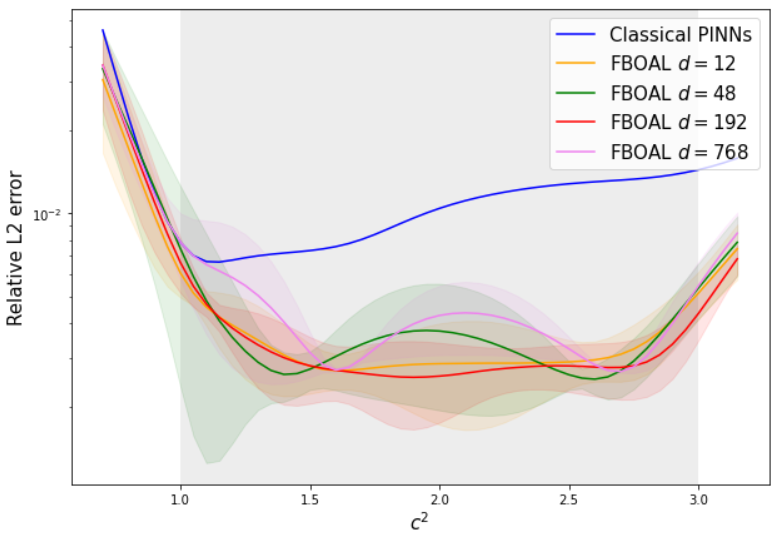} \label{wave_params_varyd}}}
    \caption{\textit{Wave equation: Performance of FBOAL when varying $m$, $k$ and $d$.} The zone in gray is the learning interval for $\nu$ (interpolation zone). The curves and shaded regions represent the geometric mean and one standard deviation of five runs.}%
    \label{wave_boal_params_vary}%
\end{figure}

\end{document}